\documentclass[lettersize,journal]{IEEEtran}
\usepackage{amsmath,amsfonts}
\usepackage{algorithmic}
\usepackage{algorithm}
\usepackage{array}
\usepackage[caption=false,font=normalsize,labelfont=sf,textfont=sf]{subfig}
\usepackage{textcomp}
\usepackage{stfloats}
\usepackage{url}
\usepackage{verbatim}
\usepackage{graphicx}
\usepackage{cite}
\usepackage{tabularray}
\usepackage{color}
\usepackage{makecell}
\usepackage{etoolbox}
\makeatletter
\patchcmd{\@makecaption}
{\scshape}
{}
{}
{}
\makeatother

\usepackage[section]{placeins}

\begin{document}

\title{CLIP-Optimized Multimodal Image Enhancement via ISP-CNN Fusion for Coal Mine IoVT under Uneven Illumination}

\author{Shuai Wang, Shihao Zhang, Jiaqi Wu, Zijian Tian, Wei Chen, Tongzhu Jin, Miaomiao Xue, Zehua Wang, Fei Richard Yu, Fellow IEEE and
	Victor C. M. Leung, life Fellow IEEE

\thanks{
	This work was supported in part by the National Natural Science Foundation of China under Grant 52274160 and Grant 51874300; in part by the National Natural Science Foundation of China-Shanxi Provincial People’s Government Coal-Based Low Carbon Joint Fund under Grant U1510115; in part by the “Jiangsu Distinguished Professor” Project in Jiangsu Province under Grant 140923070; and in part by the Fundamental Research Funds for the Central Universities Grant 2023QN1079 and Grant 2024QN11050. (Corresponding author: Wei Chen, Jiaqi Wu.)}
\thanks{Shuai Wang,Shihao Zhang,Zijian Tian,Tongzhu Jin,Miaomiao Xue are with the School of Artificial Intelligence, China University of Mining and Technology (Beijing), Beijing 100083, China.(e-mail:75172939@qq.com;286691859@qq.com;Tianzj0726@126.com;134276\\7466@qq.com;SQT2200404082@student.cumtb.edu.cn)
}
\thanks{Jiaqi Wu is with the School of Artificial Intelligence, China University of Mining and Technology (Beijing), Beijing 100083, China. The author is also with Department of Electrical and Computer Engineering, University of British Columbia, Vancouver, 2332 Main Mall Vancouver, BC Canada V6T 1Z4 (e-mail: wjq11346@student.ubc.ca)}
\thanks{Wei Chen is with the Ministry of Emergency Management Big Data Center, Beijing 100013, China. The author is also with the School of Mechanical Electronic \& Information Engineering, China University of Mining and Technology (Beijing), Beijing 100083, China. (e-mail: chenwdavior@163.com)
}
\thanks{Zehua Wang is with the Department of Electrical and Computer Engi
	neering, The University of British Columbia, Vancouver, 2332 Main Mall
	Vancouver, BC Canada V6T 1Z4 (e-mail: zwang@ece.ubc.ca).}
\thanks{F. Richard Yu is with the Department of Systems and Computer Engineering,
			Carleton University, Ottawa, ON K1S 5B6, Canada (e-mail: richard.yu@carleton.ca).}

\thanks{Victor C. M. Leung is with the Artificial Intelligence Research Institute, Shenzhen MSU-BIT University, Shenzhen 518172, China, the College of Computer Science and Software Engineering, Shenzhen University, Shenzhen 528060, China, and the  Department of Electrical and Computer Engineering, The University of British Columbia, Vancouver, BC V6T 1Z4,
			Canada (e-mail: vleung@ieee.org).}

}

\markboth{IEEE INTERNET OF THINGS JOURNAL}%
{Shell \MakeLowercase{\textit{et al.}}: A Sample Article Using IEEEtran.cls for IEEE Journals}

\IEEEpubid{0000--0000/00\$00.00~\copyright~2024 IEEE}

\maketitle

\begin{abstract}
Clear monitoring images are crucial for the safe operation of coal mine Internet of Video Things (IoVT) systems. However, low illumination and uneven brightness in underground environments significantly degrade image quality, posing challenges for enhancement methods that often rely on difficult-to-obtain paired reference images. Additionally, there is a trade-off between enhancement performance and computational efficiency on edge devices within IoVT systems.To address these issues, we propose a multimodal image enhancement method tailored for coal mine IoVT, utilizing an ISP-CNN fusion architecture optimized for uneven illumination. This two-stage strategy combines global enhancement with detail optimization, effectively improving image quality, especially in poorly lit areas. A CLIP-based multimodal iterative optimization allows for unsupervised training of the enhancement algorithm. By integrating traditional image signal processing (ISP) with convolutional neural networks (CNN), our approach reduces computational complexity while maintaining high performance, making it suitable for real-time deployment on edge devices.Experimental results demonstrate that our method effectively mitigates uneven brightness and enhances key image quality metrics, with PSNR improvements of 2.9\%-4.9\%, SSIM by 4.3\%-11.4\%, and VIF by 4.9\%-17.8\% compared to seven state-of-the-art algorithms. Simulated coal mine monitoring scenarios validate our method's ability to balance performance and computational demands, facilitating real-time enhancement and supporting safer mining operations.

\end{abstract}


\section{Introduction}
\IEEEPARstart The advent of the Internet of Video Things (IoVT) marks a pivotal advancement in real-time monitoring technologies, particularly within the context of coal mines \cite{article31}. Ensuring coal mine safety is essential for the sustainable development of the coal industry \cite{article22}\cite{article23}\cite{article24}. As Internet of Things (IoT) technologies continue to evolve, numerous IoT-based solutions have been implemented to improve safety and operational efficiency in mining, with IoT becoming increasingly integral to coal mine monitoring \cite{article30}. However, the inherently low-light and harsh conditions of underground coal mines present substantial challenges for the video quality captured by IoVT systems, compromising the reliability of data used in decision-making processes by both human operators and automated systems. Consequently, addressing these challenges through advanced image enhancement techniques is critical for optimizing the performance of IoVT systems in coal mining, ensuring accurate and timely monitoring under adverse environmental conditions.

\par For the issue, the existing image enhancement methods can be broadly classified into two categories: traditional model-based methods (e.g., gamma correction\cite{article1} Retinex method\cite{article2} and histogram equalization\cite{article3}, etc.) and deep learning-based methods\cite{article6}\cite{article8}\cite{article9}. The former belongs to physical models and adapts to different environments by manually adjusting the parameters, which has the advantages of fast speed and low computational complexity, but the labour cost is high due to the need to manually adjust the parameters several times for specific problems. The latter, the first CNN-based image enhancement algorithm proposed in the literature is known as LLNet \cite{article4}; RetinexNet \cite{article7} is divided into two parts, decomposition network and enhancement network, which decomposes the input image into  illuminant and reflective parts, \newpage \noindent and then enhances and adjusts  the two parts to achieve luminance enhancement; EnlightenGAN\cite{article10} based on Generative Adversarial Networks (GANs) is the the first successful method to introduce unpaired training into low illumination image enhancement, which can effectively enhance luminance and colour features under unsupervised conditions. However, they have limited performance to enhance the illuminance evenness problems in the low-light images captured from coal mine underground working spaces. 

\par Additionally, the traditional IoVT system adopts a centralised architecture\cite{article41}. In the system, the video camera transmits massive video data from underground spaces to the ground cloud server for video analysis and processing. The long-distance transmission of massive data presents several issues\cite{article38}\cite{article39}\cite{article40}. Primarily, it increases system latency, which is critical for time-sensitive monitoring tasks, as excessive delays can undermine the effectiveness of monitoring. Additionally, it leads to increased bandwidth pressure and reduced disaster resilience due to the extended length of transmission equipment lines. Thus, the adoption of edge computing in IoVT systems has become indispensable. Compared to advanced visual tasks, image enhancement, as a preprocessing operation, is more suitable for edge computing.Specifically, by integrating image enhancement algorithms with underground cameras, the cameras are upgraded to intelligent collection devices that combine image acquisition and smart enhancement functionalities. This enables end-to-end processing, from capturing raw low-light images to generating high-quality images. The enhanced images can be flexibly displayed on monitoring screens or used for downstream advanced visual tasks. Additionally, offloading the image enhancement task from cloud computing centers to edge devices helps reduce the computational load on cloud servers. Therefore, adopting edge computing in IoVT systems has become inevitable. 

\par Regarding to the aforementioned issues, the existing image enhancement algorithms applied in IoVT systems have the following shortcomings: 1) Difficulty in improving light unevenness. Due to the complex lighting conditions in underground coal mine, the problem of uneven illumination prevails in the captured images, however, the existing algorithms are unable to perceive the brightness, colour and other features of the local area and lack effective adaptive enhancement strategies. 2)High dependency on paired images. They relies on high-quality reference images to improve enhancement performance. However, in practical applications, it is often impossible to obtain reference images of the same scene. 3) Difficulty in avoiding artifacts. Existing unsupervised algorithms lack a clear optimisation objective during model training, resulting in artefacts often appearing in their enhanced images, such as Enlightengan and ZeroDCE. In actual coal mine surveillance works, the artefacts problem will cause a huge safety hazard; 4) Inability to edge computing. existing image enhancement algorithms have the conflict between computational complexity and enhancement performance, specifically, lightweight algorithms suitable for resource-constrained distributed computing devices have limited enhancement performance, while excellent performance of the algorithms cannot be deployed to embedded devices. They cannot be applied to coal mine IoVT system to perform highly responsive image processing.

\par To address the above problems, we propose a Multimodal Image Enhancement Method for Uneven Illumination Based on ISP-CNN Fusion Architecture specifically designed for coal mine IoVT systems. For illuminance evenness problem, we propose a two-stage image enhancement strategy, where first the image enhancement module is used to increase the global luminance, and then the image detail processing module is used to further enhance varying-level lightness in local regions. We propose an novel optimization approach based on multimodal contrast learning to realize unsupervised learning. It cyclically optimizes prompts to establish the dependencies of the texts with finer style images features, such as brightness and colour. It is also helpful to guide the subsequent brightness enhancement module to capture and enhance illuminance evenness faults. We design a fusion network of traditional image signal processing (ISP) algorithms and CNN. The ISP algorithm effectively avoids artifacts through constraint-based mapping and reduces computational complexity, making it suitable for resource-limited embedded devices. The contributions of this paper are summarized as follows:

\par (1) We propose a new paradigm that integrates image enhancement algorithms with IoVT. The proposed enhancement method achieves a balance between enhancement performance and computational complexity, and can be deployed at the edge of coal mine monitoring IoVT to enable real-time image enhancement.

\par (2) We propose a multimodal iterative optimization approach to achieve unsupervised model optimization, which guides subsequent enhancement modules in perceiving varying degrees of image style defects to address uneven brightness. Additionally, we design a two-stage enhancement network that integrates ISP and CNN to correct uneven brightness while avoiding artifacts.

\section{Related work}
\subsection{Low-light Image enhancement}
Low-light image enhancement techniques can be generally classified into two categories: traditional image enhancement methods and deep learning-based approaches. Traditional methods \cite{article1}\cite{article2}\cite{article3}, such as histogram equalization, Retinex-based algorithms, grayscale transformation techniques, and digital image signal processing algorithms applied in-camera (e.g., gamma correction, white balance, and color matrix adjustments), primarily focus on the original image. While these methods offer the benefit of being computationally lightweight, they often require extensive parameter tuning to accommodate varying scene conditions, leading to increased labor costs.
\par In recent years, the number of deep learning-based low-light image enhancement methods has significantly increased alongside the advancements in deep learning. LLNet \cite{article4}, as the pioneering deep learning approach for low-light image enhancement, enhances light intensity in low-light images through encoder stacking, yielding promising results. GLADNet \cite{article13} segments the enhancement process into two stages: light intensity evaluation and detail restoration, effectively improving low-light images through this dual-phase approach. MSRNet \cite{article5} enhances low-light images by learning an end-to-end mapping, while RetinexNet \cite{article7}, based on Retinex theory, decomposes images to achieve both enhancement and denoising, though it is prone to image distortion. EnlightenGAN \cite{article10}, employing perceptual loss functions and network attention to light intensity, generates images resembling those captured in normal lighting conditions. It is also the first low-light enhancement model to adopt unsupervised training methods. ZeroDCE \cite{article14}, another unsupervised deep learning-based enhancement network, adjusts image pixel values to achieve effective low-light image enhancement. The URetinexNet model \cite{article11} introduces a network with three learning-based modules dedicated to data-dependent initialization, efficient unfolding optimization, and user-specified illumination enhancement to improve brightness and detail. Other notable enhancement methods include those based on virtual exposure strategies \cite{article33}, recognition-integrated enhancement \cite{article34}, algorithms combining Retinex theory with Transformers \cite{article35}, implicit neural representation techniques \cite{article36}, and dual-input methods to restore detail in low-light images \cite{article37}. Collectively, these approaches have demonstrated substantial improvements in low-light image quality.
\par CLIP-LIT \cite{article15} is an unsupervised image enhancement algorithm that leverages the unsupervised learning capabilities of CLIP. While this approach mitigates the challenge of acquiring paired image datasets, it is susceptible to generating artifacts due to the absence of labeled supervision. Furthermore, the feature-awareness of current algorithms is generally constrained, limiting their ability to effectively capture local image details and address issues of localized illumination inhomogeneity.
\par We propose a novel luminance enhancement algorithm based on the CLIP contrastive learning framework, capable of unsupervised learning. This algorithm captures and adaptively enhances luminance features in heterogeneous regions without altering the original semantic information, thereby effectively addressing luminance inhomogeneity and achieving significant improvements in image luminance enhancement.
\subsection{CLIP-based visual works}
CLIP (Contrastive Language-Image Pre-Training) is a multimodal pre-training model grounded in contrastive learning. By leveraging 400 million image-text pairs, CLIP learns generalized visual and linguistic features, enabling zero-shot image classification based on detailed input cues. Its robust feature-awareness through contrastive learning and outstanding zero-shot generalization capabilities have made CLIP a popular choice in various computer vision tasks, consistently delivering impressive results.
\par A study in the literature \cite{article16} frames video motion recognition as video-text retrieval by encoding both the generated text and keyframe images using CLIP's pre-trained text and image encoders, and then computing the similarity between the two. This method effectively captures the semantic information of labels and facilitates zero-shot knowledge transfer. Another study \cite{article17} improved text encoding performance by replacing the HERO model's text encoder with CLIP's, achieving superior results on the video language understanding task benchmark. Additionally, research \cite{article18} leveraged CLIP-based models for pre-training video-text illustration logic, learning image features related to text semantics, and fine-tuning video subtitle generation based on these features, yielding excellent results. In \cite{article15}, CLIP was applied to the domain of low-light image enhancement, where a frozen CLIP encoder was used to generate cues related to normal-light images, guiding the enhancement network. While this approach effectively addresses defective features in localized regions, it is limited in the extent of enhancement and prone to artifact generation.
\par To tackle these challenges, we propose a CLIP-based low-light image enhancement algorithm. This algorithm leverages CLIP's robust feature perception capabilities to discern varying luminance levels in local detail regions. Furthermore, we introduce a novel luminance enhancement unit designed to adaptively perform pixel-level enhancement without introducing semantic inaccuracies, thereby effectively addressing the issue of uneven luminance.

\section{Methodology}
\subsection{Overview}
\begin{figure}[H]
	\centering
	\includegraphics[width=3.5in]{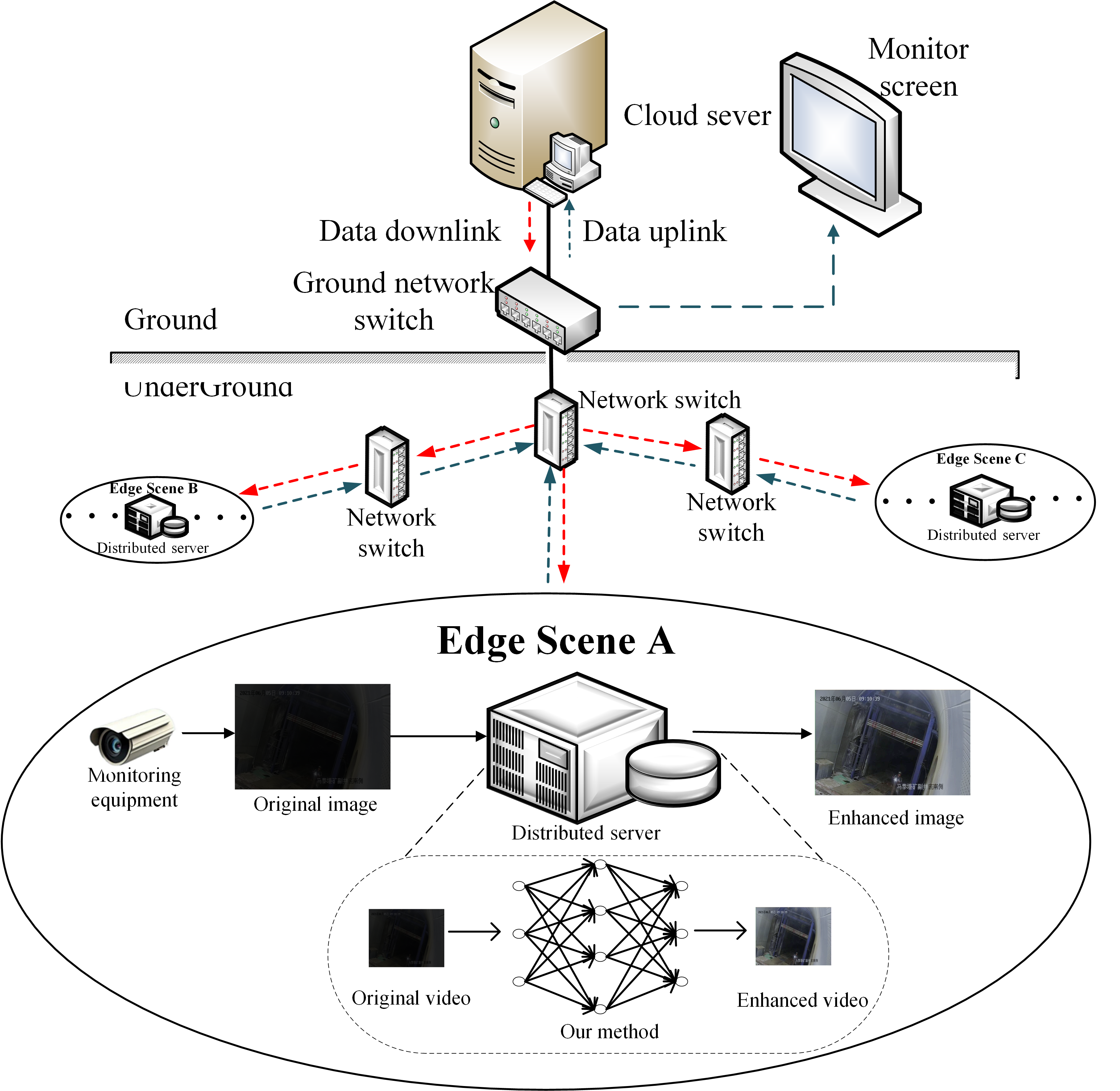}
	\caption{The architecture of the coal mine distributed IoVT system, utilizing a brightness disparity enhancement method, is depicted as follows. After the original video is captured by the monitoring equipment, the system performs light intensity enhancement within the distributed IoVT framework. The enhanced video is then transmitted to the cloud server via data transmission platforms and switches. On the cloud server, the images are further processed for advanced tasks. The red arrow indicates the data download flow, while the green arrow signifies the data upload direction.}
	\label{edge}
\end{figure}

The architecture of our distributed Internet of Video Things (IoVT) system is illustrated in Fig. \ref{edge}. Our image enhancement method is implemented on the distributed server, where it processes video streams captured by the surveillance devices.
\par Our multimodal unsupervised optimization approach is divided into three stages: the linguistic-image pairing stage, the image enhancement stage, and the cue word refinement stage, as illustrated in Fig. 2. In the linguistic-image pairing stage, we freeze the CLIP encoder to encode both the low-light image and the normal-light image, utilizing linguistic-image loss to establish dependencies between the image markers, which are connected in the latent space. In the image enhancement stage, the normal-light image encoding drives the enhancement network to improve the low-light image, resulting in an enhanced image. Finally, in the cue word refinement stage, we fine-tune the parameters of the image enhancement network to address issues of distortion and blurring that may arise during enhancement.
\begin{figure}[H]
	\centering
	\includegraphics[width=3.2in]{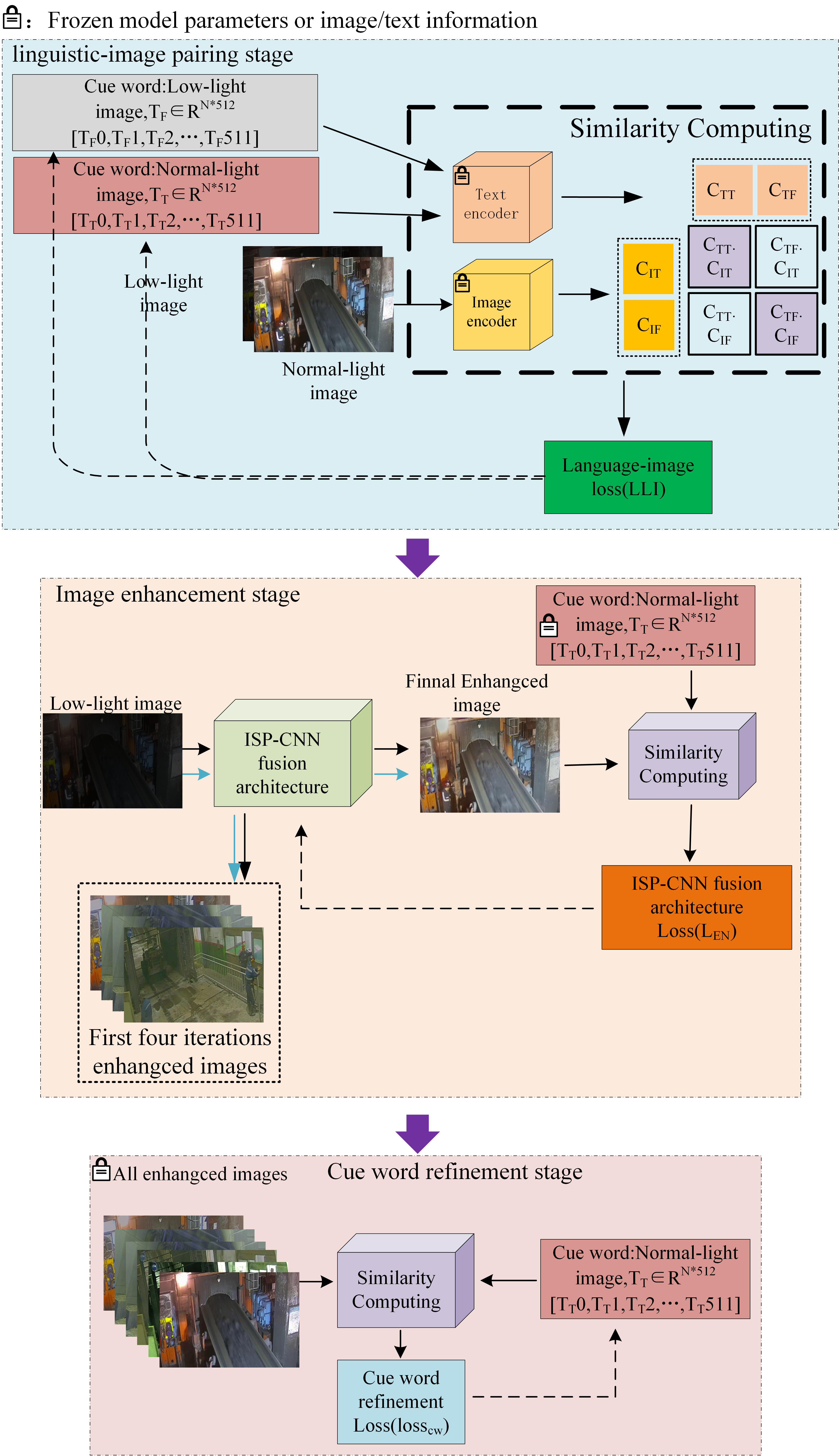}
	\caption{Structure of our algorithm. The linguistic-image pairing stage aims to optimize T$_{T}$, enabling it to differentiate between low-light and normal-light images. In the image enhancement stage, the optimized cue T$_{T}$ guides the luminance enhancement unit to perform image enhancement. Furthermore, all enhanced images generated in the previous stage are utilized to refine T$_{T}$, enhancing its ability to perceive luminance characteristics in local regions. Purple arrows indicate the computational flow during model training, dashed lines represent the optimization objectives for each loss function, and blue arrows depict the data flow during the inference stage.}
	\label{train}
\end{figure}

\subsection{CLIP Linguistic-image Pairing Stage}
The objective of the linguistic-image pairing stage is to establish a relationship between language and image information. The complete process is illustrated in Figure 2.
Initially, the low-light image (I$_{F}$) and the normal-light image (I$_{T}$) are input into the frozen CLIP image encoder (IE) to generate codes C$_{IF}$ and C$_{IT}$, respectively. Considering image quality as a binary classification problem, we introduce two randomly generated cues, "low-light image" and "normal-light image" (T$_{F}$ and T$_{T}$), into the frozen CLIP text encoder (TE), producing C$_{TF}$ and C$_{TT}$. We then design a contrastive learning-based language-image loss (LLI) algorithm to optimize T$_{FT}$ and T$_{TT}$, aligning them with the target texts (T$_{FT}$ and T$_{TT}$) that encapsulate dependencies with their corresponding images, specifically reflecting the luminance characteristics of the images.

\begin{equation}
	\label{eq1}
	{L_{LI}} = \left\{ \begin{array}{l}
		- x*\log (g),x = 1\\
		(1 - x)*\log (1 - g),x = 0
	\end{array} \right.\text{,}
\end{equation}

\begin{equation}
	\label{eq2}
	g = \frac{{{e^{({{\rm{E}}_{\rm{I}}}(i) \cdot {{\rm{E}}_{\rm{T}}}({T_T}))}}}}{{{e^{({{\rm{E}}_{\rm{I}}}(i) \cdot {{\rm{E}}_{\rm{T}}}({T_T}))}} + {e^{({{\rm{E}}_{\rm{I}}}(i) \cdot {{\rm{E}}_{\rm{T}}}({T_F}))}}}}\text{.}
\end{equation}
\par Subsequently, we optimize the hyperparameter generation network, incorporating an image detail processing module, based on the concept of contrastive learning using the loss from the image enhancement network (L$_{ehc}$).

\begin{equation}
	\label{eq3}
	{L_{ehc}} =  - \ln \frac{{{e^{({{\rm{E}}_{\rm{I}}}({I_{en}}) \cdot {{\rm{E}}_{\rm{T}}}({{\rm{T}}_{{\rm{TT}}}}))}}}}{{{e^{({{\rm{E}}_{\rm{I}}}({I_{en}}) \cdot {{\rm{E}}_{\rm{T}}}({{\rm{T}}_T}))}} + {e^{({{\rm{E}}_{\rm{I}}}({I_{en}}) \cdot {{\rm{E}}_{\rm{T}}}({{\rm{T}}_F}))}}}}\text{.}
\end{equation}
\par Where I$_{en}$ denotes the enhanced image. To minimize L$_{ehc}$, we continuously optimize the image enhancement method until I$_{en}$ becomes highly correlated with T$_{TT}$ in the latent space, thereby improving both luminance and color characteristics.
\par 
Finally, in addition to I$_{en}$, our image enhancement network generates three enhanced images with varying degrees of enhancement while preserving consistent semantic information. These images are used to further refine T$_{TT}$, shifting the focus towards luminance and color features rather than semantic content. To facilitate this refinement, we design a cue word refinement loss (loss$_{cw}$), based on marginal ordering loss, where multiple pseudo-supervised samples provide stronger regularization. The loss$_{cw}$ equation is as follows:

\begin{equation}
	\label{eq4}
	los{s_{cw}} = \sum\limits_{i = 0}^5 {\max (0,{S_i})}\text{,}
\end{equation}

\begin{equation}
	\label{eq5}
	 \begin{array}{l}
		{S_0} = {P_0} - (r({{\rm{I}}_T}) - r({{\rm{I}}_F})),\\
		{S_1} = {P_1} - (r({{\rm{I}}_T}) - r({{\rm{I}}_F})),\\
		{S_2} = {P_2} - (r({{\rm{I}}_{en}}) - r({{\rm{I}}_{{\rm{en3}}}})),\\
		{S_3} = {P_2} - (r({{\rm{I}}_{{\rm{en3}}}}) - r({{\rm{I}}_{{\rm{en2}}}})),\\
		{S_4} = {P_2} - (r({{\rm{I}}_{{\rm{en2}}}}) - r({{\rm{I}}_{{\rm{en1}}}})),\\
		{S_5} = {P_2} - (r({{\rm{I}}_{{\rm{en1}}}}) - r({{\rm{I}}_{{\rm{en0}}}}))\text{.}
	\end{array}
\end{equation}
\par In the context of an expression, r(i) represents the degree of correlation between T$_{TT}$ and the input image i, where r(i) $\in$ [0,1]. P$_{0}$ denotes the target gap between r(I$_{T}$) and r(I$_{F}$), with P$_{0}$ set to 0.9 to ensure that the correlation between T$_{TT}$ and the normal-light image is significantly higher than that with the low-light image. I$_{en0}$, I$_{en1}$, I$_{en2}$, and I$_{en3}$ represent the results of the first four iterations of our luminance enhancement module. These iterations exhibit progressively weaker enhancement compared to I$_{en}$, while maintaining the same semantic information as I$_{en}$. Thus, we set P$_{2}$ to 0.3 to maximize T$_{TT}$'s focus on luminance and color features. Since the content of I$_{T}$ and I$_{en}$ differs, P$_{1}$ is set to 0.2, which is lower than P$_{2}$, making it easier for T$_{TT}$ to capture the dependencies of luminance and color features rather than semantic information.

\subsection{Image enhancement stage} 
Given the superior feature-capturing capabilities of multimodal image enhancement methods, we designed an ISP-CNN fusion architecture, as depicted in Fig. 3, which employs a reduced number of channels. This design not only ensures the lightweight nature of the ISP-CNN fusion architecture, enhancing both efficiency and performance, but also significantly improves image quality. The architecture is primarily composed of two key components: the image enhancement module and the image detail processing module, which we will explain in detail below.

\begin{figure}[H]
	\centering
	\includegraphics[width=3.2in]{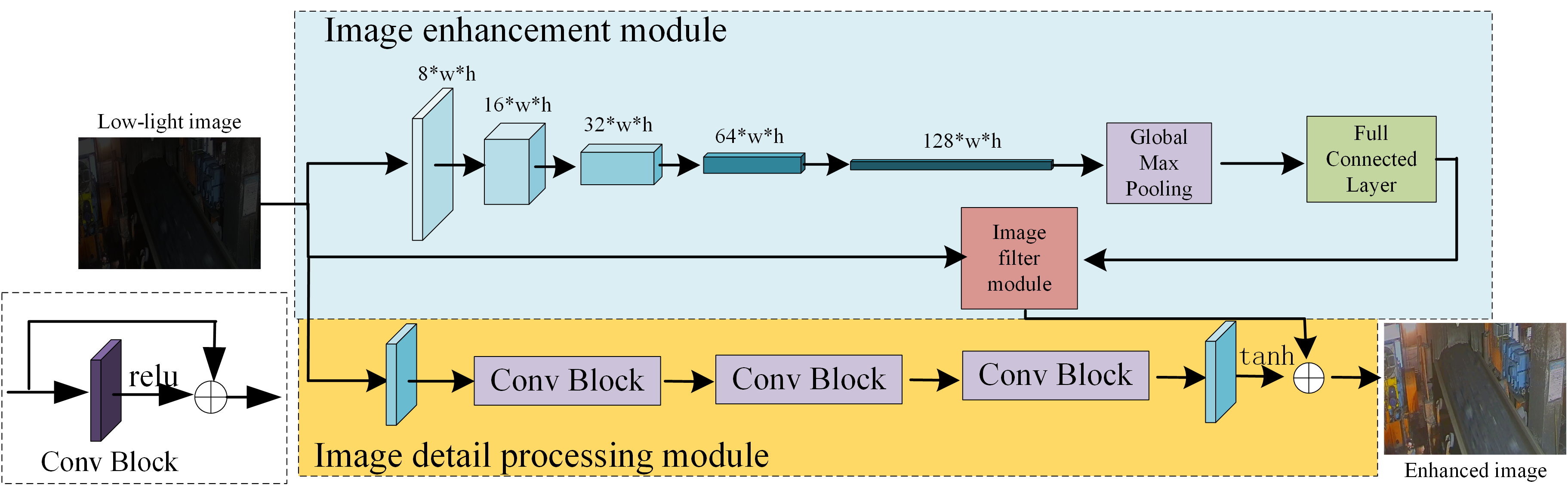}
	\caption{The structure of our ISP-CNN fusion architecture is as follows. In the image enhancement module, hyperparameters are generated through convolutional mapping, while in the image detail processing module, a mapping map is produced to further refine and correct the output from the image enhancement module.}
	\label{enh}
\end{figure}
\par {\bf{Image Enhancement Module.}}The image enhancement module is designed to improve low-illumination images and operates in two key steps: first, generating hyperparameters based on image features, and second, applying these hyperparameters within a function in the image processing unit to achieve image enhancement. Consequently, this module can be divided into two main units: the encoder unit and the image processing unit.
\par The objective of the encoder section is to leverage latent features to compute the necessary parameters for image processing. To achieve this, we employed a visual encoder consisting of five convolutional layers, each with a kernel size of 3 and a stride of 1. The number of channels in each layer doubles progressively, starting with 8 channels in the first layer and reaching 128 channels in the final layer. Each convolutional layer is followed by max pooling (with a kernel size of 3 and a stride of 2), while the final layer is succeeded by global max pooling, producing an output of 1x1x128. The pooled result is then parameterized using a fully connected layer. These parameters, generated by the visual encoder, are subsequently applied by the image processing unit to enhance the image.
\par The proposed image processing section consists of four differentiable filters with adjustable hyperparameters: white balance (WB), gamma correction, contrast, and sharpening. Standard color and hue adjustments, such as WB, gamma correction, and contrast, function as pixel-by-pixel filters. Accordingly, the filters employed in this section are categorized into pixel-level filters and an image sharpening module.
\par The pixel-weighted filter maps the input pixel value In$_{i}$ = (r$_{i}$, g$_{i}$, b$_{i}$) to the output pixel value O = (r$_{o}$, g$_{o}$, b$_{o}$), where (r, g, b) represent the values of the red, green, and blue color channels, respectively. The mapping functions for the four pixel filters are provided below. Both WB and Gamma perform simple multiplication and power transformations. Clearly, their mapping functions are localized, operating directly on the input image and its corresponding parameters.
\par White balance:
\begin{equation}
	\label{eq6}
	{O_{wb}} = f({w_r}*{r_i},{w_g}*{g_i},{w_b}*{b_i}).
\end{equation}
\par Gamma correction:
\begin{equation}
	\label{eq7}
	{O_{gamma}} = I{n_i}^{gamma}\text{.}
\end{equation}
\par Contrast:
\begin{equation}
	\label{eq8}
	\begin{array}{l}
		{O_{contrast}} = \alpha *En(I{n_i}) + (1 - \alpha )*I{n_i}\text{,} \vspace{1.5ex}  \\ 
		EnL(I{n_i}) = \frac{1}{2}*(1 - \cos (\pi *(0.27*{r_i} + 0.67*{g_i} +\\ 0.06*{b_i})))\text{,} \vspace{1.5ex} \\
		En(I{n_i}) = I{n_i}*\frac{{EnL(I{n_i})}}{{0.27*{r_i} + 0.67*{g_i} + 0.06*{b_i}}}\text{.}
	\end{array}
\end{equation}
\par Sharpening filter. Image sharpening highlights image details:
\begin{equation}
	\label{eq9}
	{O_{sharpen}} = I{n_i} + \lambda *({P_i} - Gauss(I{n_i}))\text{.}
\end{equation}
\par Where In$_{i}$ represents the input image, Gauss (In$_{i}$) denotes the Gaussian filter, and $\lambda$ is a positive scaling factor. The sharpening effect can be adjusted by optimizing $\lambda$, enhancing object detection performance.
\par {\bf{Image Detail Processing Module.}}We designed the image detail processing module to address the minor distortions that can arise from the overall image enhancement process. Although the image enhancement module incorporates sharpening, contrast adjustment, and other techniques to refine image details, it inevitably introduces some degree of distortion. To mitigate this, we employ a feature mapping network composed of three residual modules (each with a kernel size of 3 and a stride of 1), as illustrated in Fig. 3. This module is capable of learning and mapping image detail information on a pixel-by-pixel basis. When integrated with the image enhancement module, it effectively restores fine image details. The first layer maps the three input channels into 32 channels, with each subsequent residual module also maintaining 32 channels. The final layer maps the 32 channels back to the original three channels. Each convolution operation is followed by a batch normalization (BN) layer and a ReLU activation layer, while the last layer uses a Tanh activation function. The final output is then combined with the output of the image enhancement module to produce the enhanced image.

\section{Experiment}
\subsection{Experimental Settings}
In this paper, we utilize the coal mine dataset as the primary dataset, which consists of 600 low-light/normal-light image pairs, each with dimensions of 400×600. Additionally, we incorporate the public datasets SICE and LOL to further evaluate the algorithm's performance. The LOL dataset includes 500 low-light/normal-light image pairs, each with dimensions of 400×600, while the SICE dataset comprises 200 such image pairs, also with dimensions of 400×600. 
 The specific experimental environments and main hyperparameters are outlined in Table I. In our comparative experiments, we selected several state-of-the-art low-light image enhancement algorithms, including RetinexNet \cite{article7}, URetinexNet \cite{article11}, EnlightenGAN \cite{article10}, Clip-LIT \cite{article15}, SCI \cite{article21}, ZERO-DCE \cite{article14}, and IAT \cite{article20}, and conducted comparison experiments on both the coal mine dataset and the public datasets.

\par We selected full-reference image quality assessment (IQA) metrics, including Visual Information Fidelity (VIF), Peak Signal-to-Noise Ratio (PSNR), and Structural Similarity (SSIM), along with the state-of-the-art non-reference metric, Multiscale Image Quality Transformer (MUSIQ), to quantitatively evaluate image quality.
\par The VIF equation:
\begin{equation}
	\label{eq10}
	VIF = \frac{{\sum\limits_{i = 1}^n {I({X_i};{Y_i})} }}{{\sum\limits_{i = 1}^n {I({X_i};{Z_i})} }}\text{.}
\end{equation}
\par Where X$_{i}$ and Y$_{i}$ represent the information subbands of the reference image and the distorted image, respectively, Z$_{i}$ denotes the output of the human visual system, I represents the mutual information, and Z$_{i}$ indicates the number of subbands.
\par SSIM equation:
\begin{equation}
	\label{eq12}
	SSIM(x,y) = \frac{{(2{\mu _x}{\mu _y} + {c_1})(2{\sigma _{xy}} + {c_2})}}{{({\mu _x}^2 + {\mu _y}^2 + {c_1})({\sigma _x}^2 + {\sigma _y}^2 + {c_2})}}\text{.}
\end{equation}
\par Where x, y are the enhanced image and the target image, $\mu$$_{x}$, $\mu$$_{y}$ are the image means, $\sigma$$_{x}$, $\sigma$$_{y}$ are the image standard deviation, $\sigma$$_{xy}$ is the image covariance, c$_{1}$, c$_{2}$ are the auxiliary constants.
\par Equation for MUSIQ:
\begin{equation}
	\label{eq13}
	Q = \frac{1}{N}\sum\limits_{i = 1}^n {\sigma ({w_q}^T{h_i})}\text{.}
\end{equation}
\par Where Q is the quality fraction of the image, N is the number of pixels, $\sigma$ is the sigmoid function, w$_{q}$ is the weight vector of the image quality, and h$_{i}$ is the feature vector of the i-th pixel encoded by the Transformer.

\subsection{Ablation experiments}
We conducted ablation experiments using the SICE dataset to evaluate the effectiveness of several key components.

\par First, we assessed the feature extraction capabilities of CLIP. Specifically, we examined the impact of labeled images with varying brightness levels or differing semantic content. As shown in Fig. 4, labeled images with different luminance levels significantly affected the enhancement performance of our algorithm. In contrast, images with differing semantic content but consistent luminance did not present the same issue. This highlights the improved cue words' strong ability to perceive luminance features and demonstrates the effectiveness of CLIP's advanced feature extraction capabilities in our method.

\begin{figure}[!h]
	\centering
	\includegraphics[width=3in]{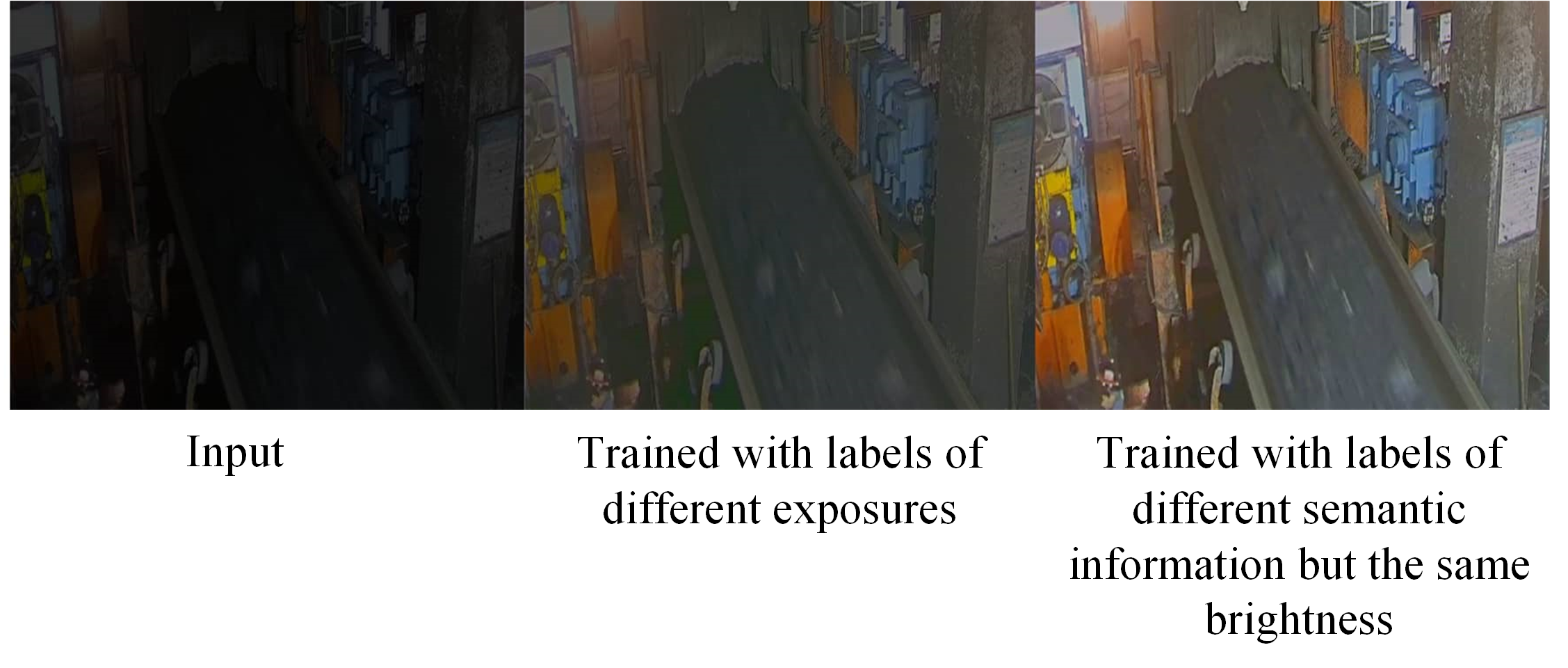}
	\caption{Algorithm performance based on images with different training labels is illustrated. From left to right, the images show the input, the inference results trained with labels of varying exposures, and the inference results trained with labels of differing semantic information but consistent brightness. A significant difference in enhancement effect is observed between models trained with different label images.}
	\label{unsp}
\end{figure}

\par Effectiveness of the cue word refinement stage: As shown in Fig. 5, when the cue word refinement stage is omitted, our algorithm still achieves overall luminance enhancement. However, the luminance features in local detail regions are not effectively improved, demonstrating the importance of the cue word refinement strategy. This strategy significantly enhances the algorithm's luminance performance, ensuring that the refined cues focus more on the luminance of heterogeneous regions. Furthermore, Fig. 5 shows how the distribution of the final image detail processing module reinforces the effectiveness of cue refinement. By refining the cues, the detail processing module is guided to accurately capture luminance features, providing a solid foundation for subsequent adaptive enhancement.

\begin{figure}[H]
	\centering
	\includegraphics[width=3in]{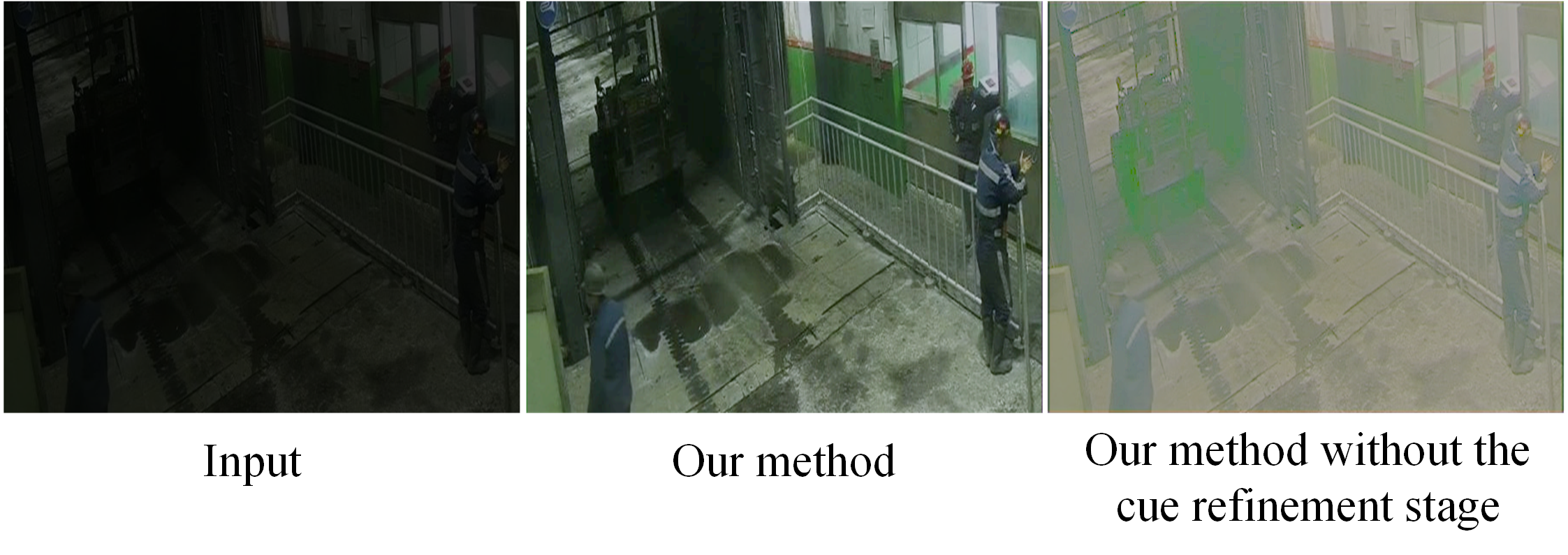}
	\caption{Enhancement results of our algorithm with and without the cue word refinement stage: The images, in sequence, are the input image, the algorithm-enhanced image, the algorithm-enhanced image without the cue refinement stage, and the reference image. The version of the algorithm without the cue refinement stage exhibits limited brightness enhancement in localized regions.}
	\label{cw}
\end{figure}
\par The results above demonstrate that the proposed cue word refinement is the key technology for addressing luminance uniformity in our method.

\par Quantitative Comparison: To further validate this conclusion, we trained the model using different strategies and conducted a quantitative analysis of the enhancement results, as shown in Table I.
\definecolor{Cerulean}{rgb}{0.058,0.619,0.835}
\begin{table}[H]
	\scriptsize
	\centering
	\caption{Effect of different learning approaches: In each IQA metric, the best results are highlighted in bold font.}
	\begin{tblr}{
			hline{1-2,5} = {-}{},
		}
		Training
		approach                                       & PSNR & SSIM & VIF      & MUSIQ   \\
		{
			Our approach without~\\cue word refinement stage}     & 14.21 & 0.452 & 
		0.5629 & 
		49.33 \\
		{Training with labeled images \\with different exposures} & 11.06 & 0.251 & 0.4674    & 33.42    \\
		Our approach                                              & $\bf{16.35}$ & \bf{0.595} & \bf{0.6770}    & \bf{65.70}    
	\end{tblr}
\end{table}
\par From Table I, it is evident that training with differently exposed images or omitting the cue word refinement phase leads to a decline in performance. Notably, the algorithm experiences the greatest performance degradation when trained with differently exposed images.
\subsection{Comparative experiments}
\par Comparative experiments on the coal mine dataset:
To highlight the practicality and superiority of our approach, we conducted comparative experiments using various image enhancement methods on the coal mine dataset, as illustrated in Fig. 6.
\begin{figure*}[h]
	\centering
	\includegraphics[width=7in]{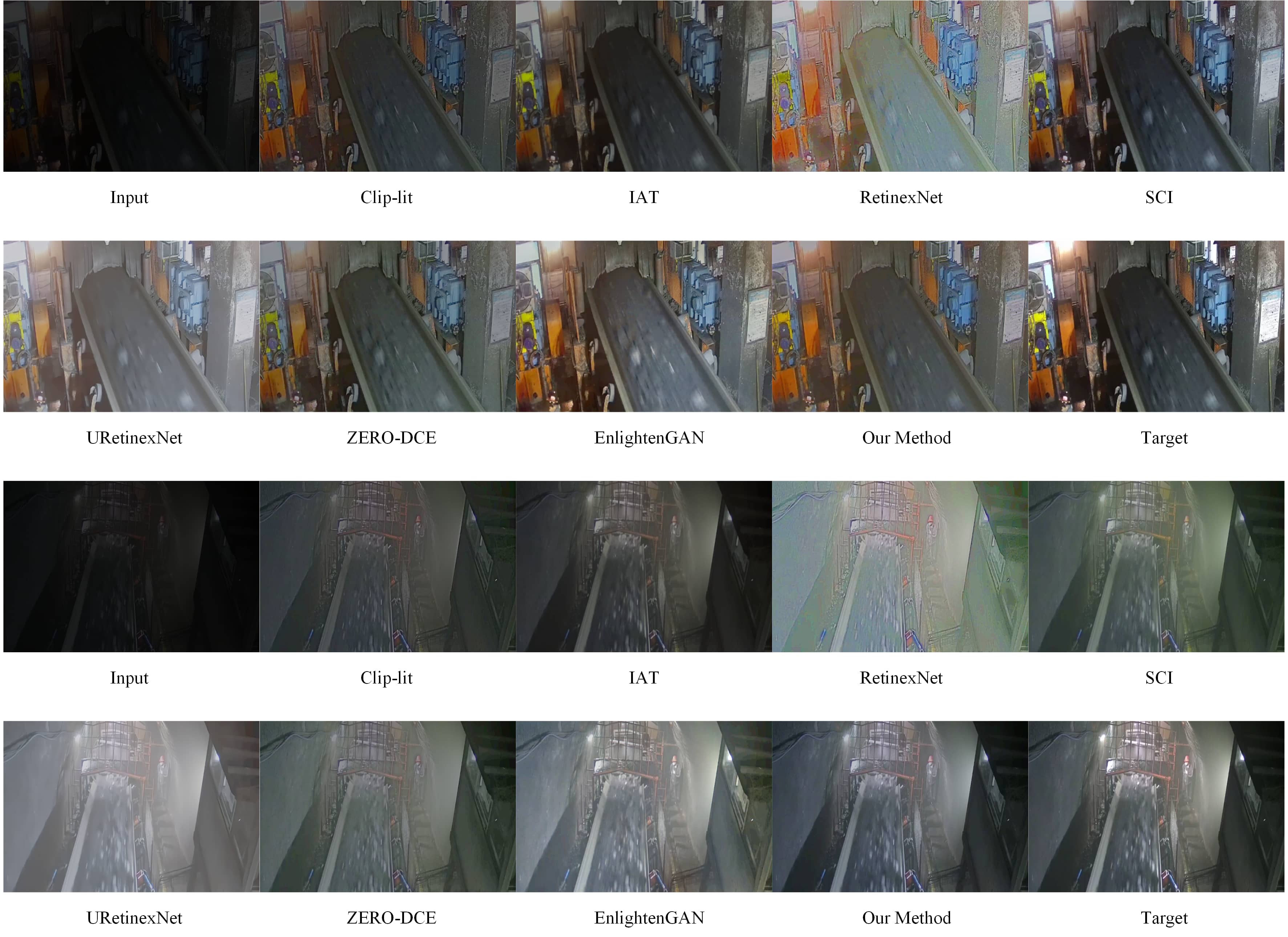}
	\caption{Comparison of image enhancement results between our algorithm and other algorithms using samples from the coal mine dataset. It is evident that our algorithm provides superior visual enhancement.}
	\label{ccmp}
\end{figure*}
\begin{figure*}[t]
	\centering
	\includegraphics[width=7in]{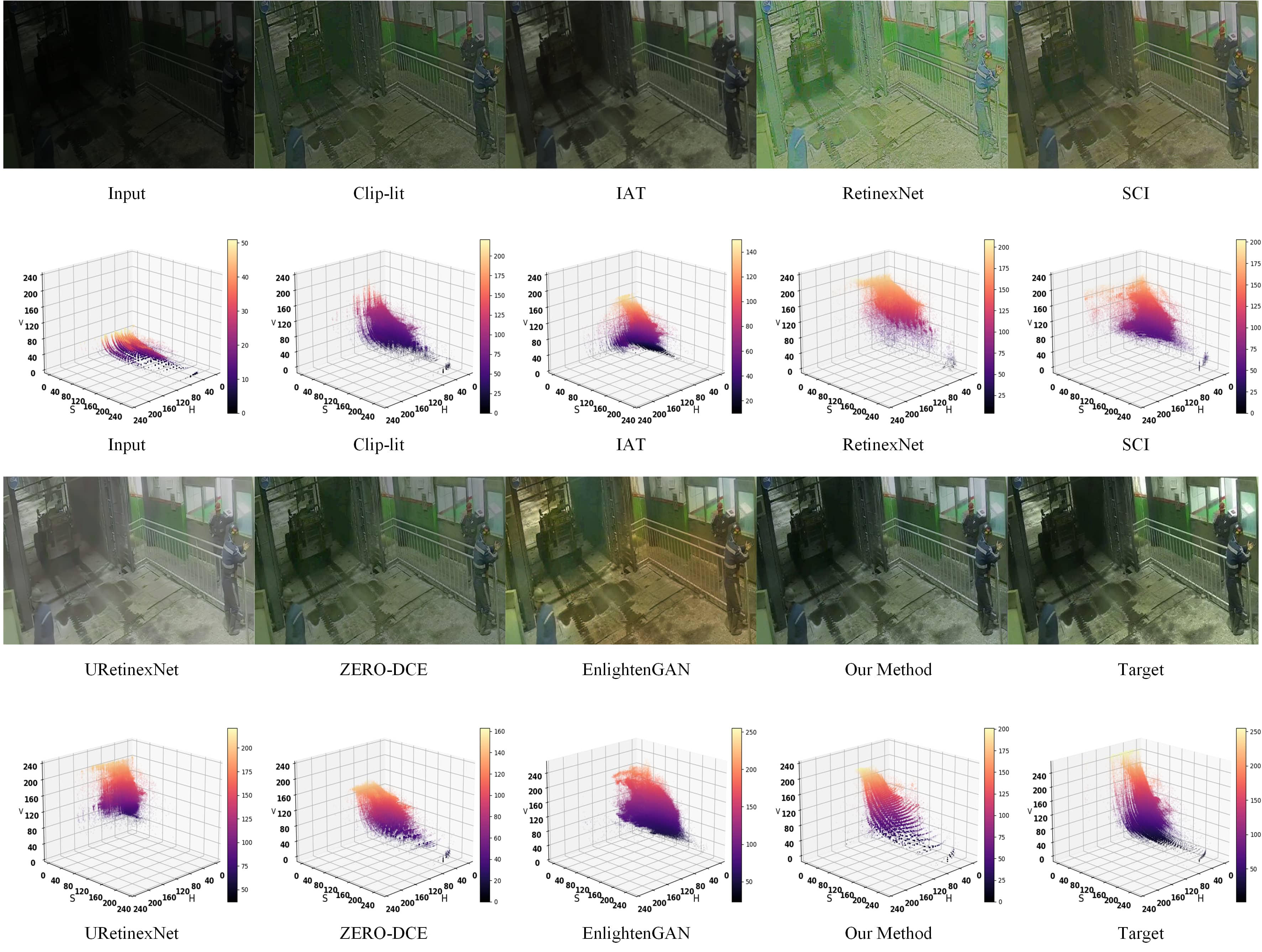}
	\caption{Comparison of HSV maps between the results of our algorithm and those of other algorithms. The HSV plots indicate that the output of our algorithm more closely aligns with the target.}
	\label{hsv}
\end{figure*}
\par As shown in Fig. 6 and Fig. 7, Clip-LIT and IAT \cite{article20} have a limited effect on overall image brightness, while the RetinexNet-enhanced image exhibits significant color distortion. Although SCI \cite{article21} and URetinexNet perform well in enhancing brightness, they still result in a degree of color distortion. ZERO-DCE and EnlightenGAN are less effective in processing fine details, often leading to issues like ghosting and false information. In contrast, our algorithm not only significantly improves image brightness and successfully addresses brightness inhomogeneity, but also accurately captures and adaptively enhances the brightness characteristics of heterogeneous regions. Furthermore, it effectively preserves and enhances color features, preventing color distortion.

\par Quantitative comparison on coal mine dataset.
To further illustrate the advantages of our algorithm in terms of both performance and computational efficiency, we applied it to the coal mine dataset and compared the IQA metrics of the inference results with those of other algorithms. The results are presented in Table II. 

\definecolor{Cerulean}{rgb}{0.058,0.619,0.835}
\begin{table}[H]
	\scriptsize
	\centering
	\caption{Quantitative comparison of IQA metrics on the coal mine dataset: the best results are highlighted in bold font.}
	\begin{tblr}{
			hline{1-2,10} = {-}{},
		}
		Algorithm        & PSNR        & SSIM        & VIF          & MUSIQ       \\
		RetinexNet\cite{article7}    & 11.37       & 0.279       & 0.3813       & 59.81       \\
		URetinex-Net\cite{article11} & 12.78       & 0.456       & 0.4080       & 
		63.48
		\\
		EnlightenGAN\cite{article10} & 14.48       & 0.486       & 
		0.6452
		& 59.35       \\
		CLIP-LIT\cite{article15}     & 12.97       & 0.494       & 0.4132       & 57.36       \\
		SCI\cite{article21}              & 
		15.56
		& 
		0.534
		& 0.5740       & 60.09       \\
		ZERO-DCE\cite{article14}         & 14.79       & 0.450       & 0.5290       & 60.29       \\
		IAT\cite{article20}              & 13.64       & 0.376       & 0.4213       & 53.56       \\
		Our algorithm    & 
		\bf{16.35}
		& 
		\bf{0.595}
		& 
		\bf{0.6770}
		& 
		\bf{65.72}
		
	\end{tblr}
\end{table}
\par In Table II, our algorithm outperforms the comparison algorithms across all IQA metrics, demonstrating superior image enhancement performance in terms of brightness enhancement, color enhancement, and preservation of original sensory information (PSNR: 4.8\%, SSIM: 11.4\%, VIF: 4.9\%).

\par Comparative experiments on public datasets:
To provide a more comprehensive evaluation of our method's performance, we conducted additional comparative experiments on public datasets, including LOL and SICE. The inference results are shown in Fig. 8.

\begin{figure}[H]
	\centering
	\includegraphics[width=3.5in]{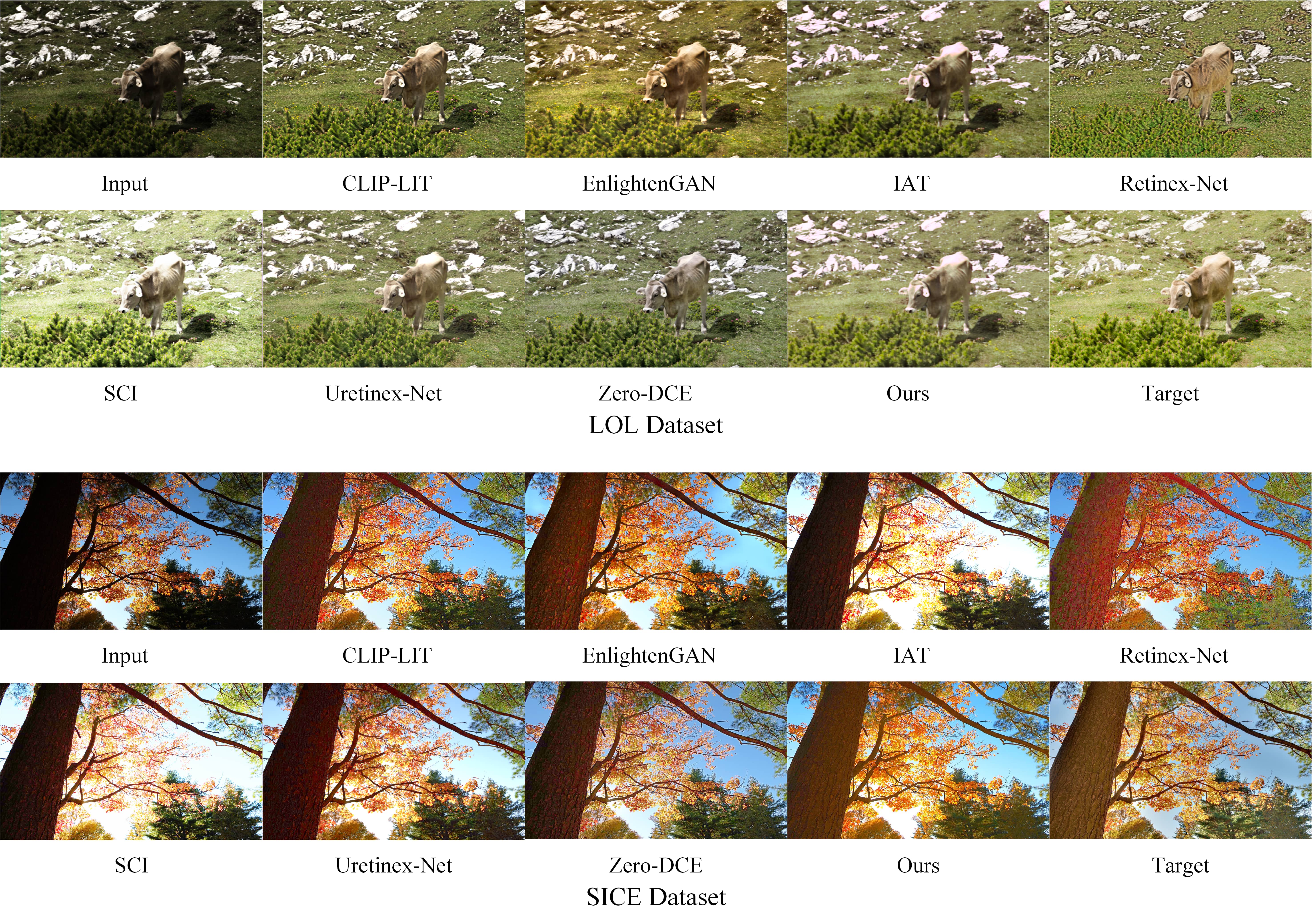}
	\caption{Comparison of image enhancement results between our algorithm and other algorithms using samples from the LOL and SICE datasets. It is evident that the images enhanced by our algorithm exhibit superior visual quality.}
	\label{cmpg}
\end{figure}

\par As shown in Fig. 8, the Clip-LIT, SCI, and IAT algorithms did not significantly enhance image brightness, while the RetinexNet-enhanced image exhibited severe color distortion. The remaining algorithms also introduced varying degrees of color imbalance during the enhancement process. Although Zero-DCE and EnlightenGAN produced relatively good enhancement results, both algorithms introduced ghosting and false information. In contrast, our proposed algorithm significantly improves overall image brightness while excelling in color enhancement, effectively mitigating the issue of color distortion.

\par Quantitative comparison on public datasets.
To quantitatively assess the performance of our algorithm, we applied it to the LOL dataset and calculated the IQA metrics for the resulting inferences. The results are presented in Table III.
\definecolor{Lochmara}{rgb}{0,0.439,0.752}
\begin{table}[H]
	\scriptsize
	\centering
	\caption{Quantitative comparison of IQA metrics on the LOL dataset: the best results are highlighted in bold font.}
	\begin{tblr}{
			hline{1-2,10} = {-}{},
		}
		Algorithm        & PSNR        & SSIM        & VIF          & MUSIQ       \\
		RetinexNet\cite{article7}    & 16.28       & 0.513       & 0.1665       & 57.28       \\
		URetinex-Net\cite{article11} & 
		21.49
		& 
		0.803
		& 
		0.5191
		& 
		\bf{70.61}
		\\
		EnlightenGAN\cite{article10} & 18.48       & 0.792       & 0.3272       & 56.24       \\
		CLIP-LIT\cite{article15}     & 16.94       & 0.768       & 0.4594       & 56.69       \\
		SCI\cite{article21}              & 20.07       & 0.772       & 0.4956       & 56.13       \\
		ZERO-DCE\cite{article14}         & 16.45       & 0.665       & 0.3304       & 55.29       \\
		IAT\cite{article20}              & 20.82       & 0.796       & 0.5012       & 63.98       \\
		Our algorithm    & 
		\bf{22.55}
		& 
		\bf{0.863}
		& 
		\bf{0.5917}
		& 
		69.05
		
	\end{tblr}
\end{table}

\par In Table III, our algorithm outperforms the comparison algorithms across all IQA metrics, demonstrating superior image enhancement performance in terms of brightness enhancement, color enhancement, and preservation of original visual information (PSNR: 4.9\%, SSIM: 7.4\%, VIF: 14.0\%).

\par `
Backlight enhancement experiment:
To further emphasize the effectiveness of our proposed method in handling unevenly illuminated images, we conducted comparative experiments using the backlit dataset, BacklitNet.

\begin{figure}[H]
	\centering
	\includegraphics[width=3.5in]{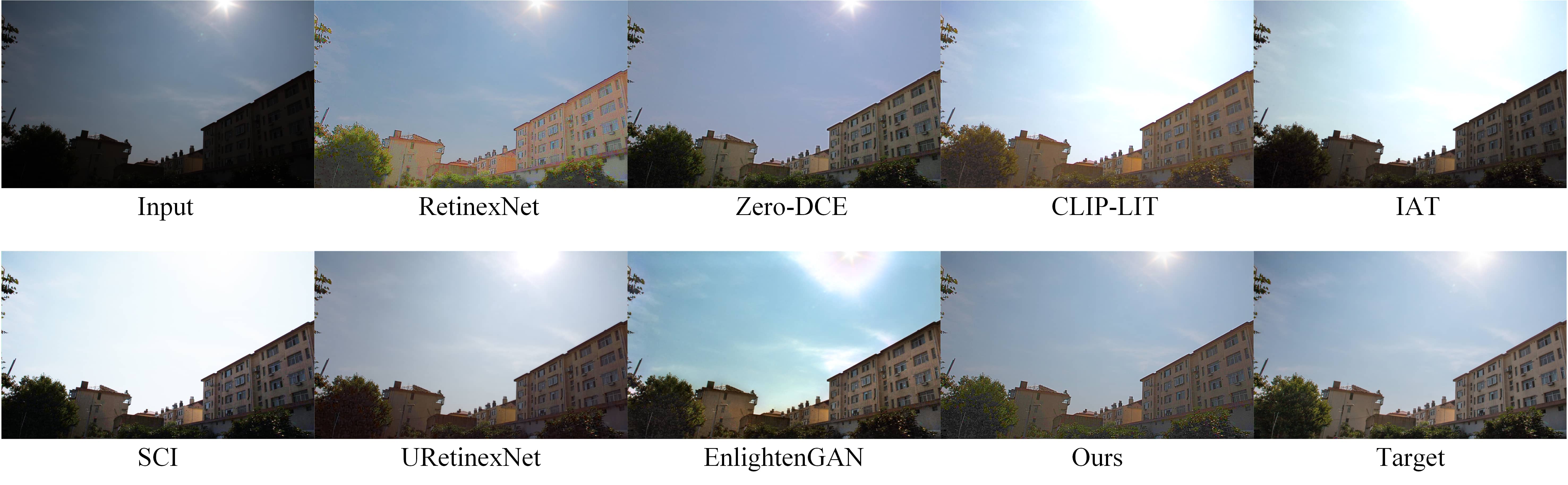}
	\caption{Comparison of image enhancement results between our algorithm and other algorithms using samples from the BacklitNet dataset.}
	\label{back}
\end{figure}
\par The experimental results reveal that EnlightenGAN and IAT algorithms tend to suffer from over-enhancement in regions of higher brightness in the input image, while algorithms such as Zero-DCE, SCI, and URetinex exhibit insufficient enhancement in areas with lower brightness. Other algorithms display varying degrees of detail distortion during processing. In contrast, our proposed algorithm demonstrates superior performance, effectively handling scenes with uneven illumination.

\subsection{Coal Mine Simulation Experiment}
\par In the simulation experiment, we selected the same algorithms used in the comparison experiment for further testing. ZERO-DCE, IAT, and our algorithm were deployed on edge servers, while the remaining algorithms were deployed on cloud servers. The enhancement results for the monitoring data are shown in Fig. 10. We conducted a quantitative analysis of the simulation results and performed a statistical analysis of the computational complexity of each algorithm, as presented in Table IV.
\begin{figure*}[h]
	\centering
	\includegraphics[width=7in]{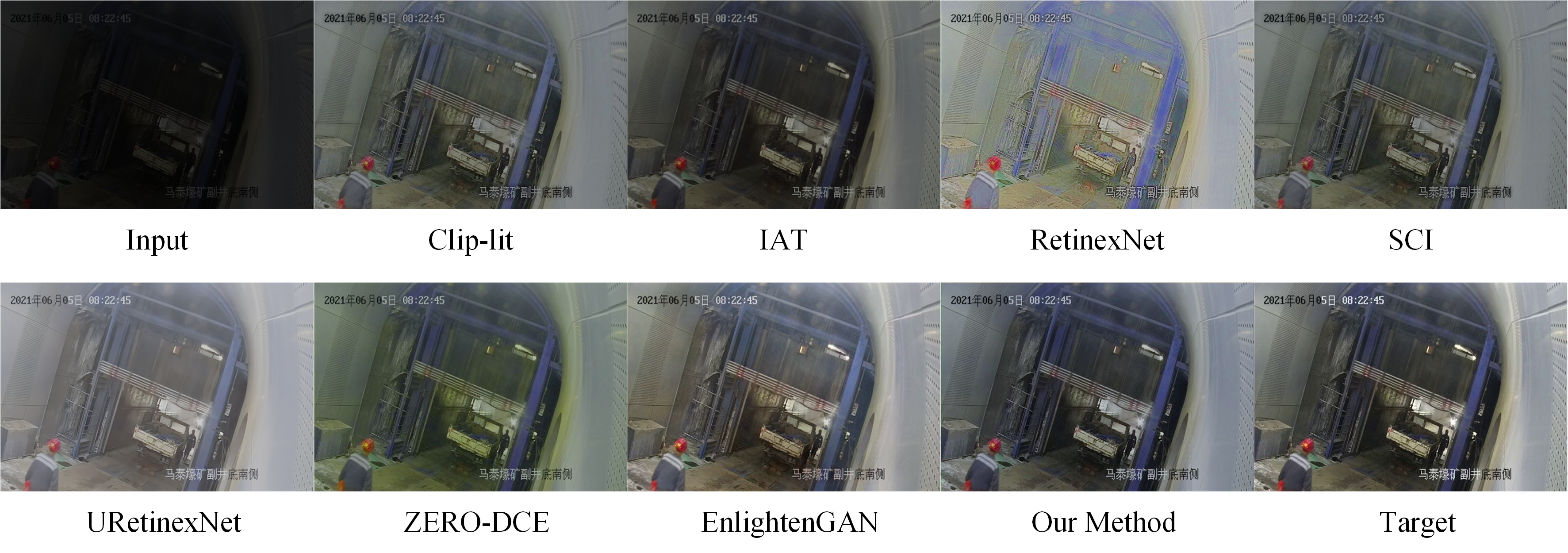}
	\caption{Comparison of data transmission delays under different computing methods: The delay in edge computing remains relatively stable regardless of the volume of transmitted data. In contrast, cloud computing experiences a significant increase in delay as the amount of transmitted data grows.}
	\label{simu}
\end{figure*}
\definecolor{Cerulean}{rgb}{0.058,0.619,0.835}
\begin{table}[!h]
	\centering
	\caption{Quantitative comparison of IQA metrics in the simulation experiment:the best results are highlighted in bold font.}
	\scalebox{0.8}{
	\begin{tblr}{
			hline{1-2,6,10} = {-}{},
		}
		{
			The computing \\approach
		}      & Algorithm        & PSNR        & SSIM        & VIF          & MUSIQ     &FLOPs  \\
		{
			Centralized cloud \\computing
		} & RetinexNet\cite{article7}    & 12.61       & 0.373       & 0.4536       & 63.45   & 73.93G    \\
		& URetinex-Net\cite{article11} & 13.42       & 0.508       & 0.5431       & 
		\bf{73.72} & 208.42G
		\\
		& EnlightenGAN\cite{article10} & 15.68       & 0.562       & 0.6972       & 65.69    &72.6G   \\
		& CLIP-LIT\cite{article15}     & 13.90       & 0.614       & 0.5716       & 61.94     &66.81G  \\
		Edge computing                        & SCI\cite{article21}              & 
		16.85
		& 
		0.673
		& 
		0.7402
		& 65.93    &41.2G   \\
		& ZERO-DCE\cite{article14}         & 16.17       & 0.627       & 0.5893       & 65.66   &19.2G    \\
		& IAT\cite{article20}              & 14.40       & 0.644       & 0.5792       & 56.19    &\bf{5.25G}   \\
		& Our algorithm    & 
		\bf{17.35}
		& 
		\bf{0.702}
		& 
		\bf{0.7534}
		& 
		67.72
		&12.1G
		
	\end{tblr}
}
\end{table}
\par In Table IV, our algorithm outperforms the comparison algorithms across all IQA metrics, demonstrating superior image enhancement performance in terms of brightness enhancement, color enhancement, and preservation of original visual information (PSNR: 2.9\%, SSIM: 4.3\%, VIF: 17.8\%).
\par From the above experiments, it is evident that our algorithm ranks second in terms of computational complexity, demonstrating a low computational cost. However, it outperforms all other algorithms in terms of performance, achieving an optimal balance between performance and computational efficiency. This balance makes our algorithm both highly effective and economical, making it well-suited for deployment in distributed IoVT.
\par We deployed the algorithms on both cloud servers and edge servers and calculated their inference speeds (FPS) separately. Our distributed deployment strategy significantly improves the overall response speed of image enhancement compared to the centralized cloud server deployment approach.
\begin{table}[H]
	\centering
	\caption{Total response speed of the algorithm across different deployment methods.}
	\setlength{\tabcolsep}{10mm}{
	\begin{tabular}{ll} 
		\hline
		Deployment position     & FPS    \\ 
		\hline
		Cloud server & 3.63   \\
		Edge servers & 15.16  \\
		\hline
	\end{tabular}
	}
\end{table}

\par As shown in Table V, the FPS (frames per second) of the algorithm deployed on cloud servers is significantly lower compared to the algorithm deployed on edge servers.

\par In Fig. 11, the increase in data volume during communication leads to higher latency, which is particularly pronounced in cloud computing methods. This increase in latency impacts the efficiency of subsequent advanced image processing methods. In contrast, latency in edge computing remains consistently low. For instance, the communication latency in cloud computing reaches 248.4 ms for the transmission of 40 images, whereas in edge computing, it is only 9.7 ms. These results indicate that the edge computing approach has a significantly lower impact on overall algorithm responsiveness, making it more suitable for real-time tasks compared to cloud computing.
\section{Conclusion}
\begin{figure}[!t]
	\centering
	\includegraphics[width=3in]{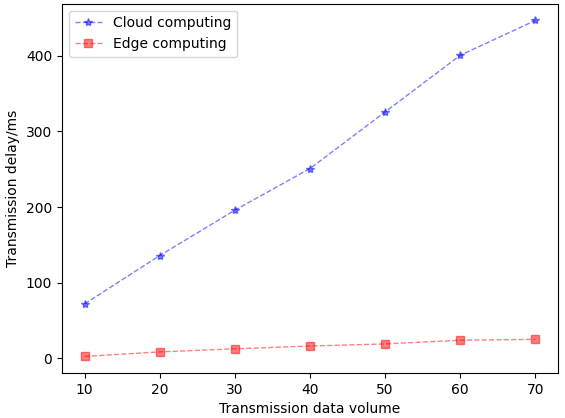}
	\caption{Comparison of data transmission delays across different computing methods: In edge computing, transmission delay is minimally impacted by the volume of data. However, in cloud computing, delay increases significantly as the volume of transmitted data grows.}
	\label{delay}
\end{figure}
To address the issues of low brightness and uneven illumination in coal mine Internet of Video Things (IoVT), we propose a multimodal image enhancement method based on an ISP-CNN fusion architecture. By incorporating the CLIP model, this method achieves multimodal unsupervised learning, eliminating the need for reference images and enhancing its applicability in real-world industrial scenarios. The method employs a cyclic optimization strategy and a two-stage feature enhancement network, enabling adaptive correction of local style feature defects, significantly improving image quality, particularly in uneven brightness regions. Moreover, the fusion of traditional ISP with CNN reduces computational complexity while maintaining excellent enhancement performance, allowing real-time image enhancement on edge devices in coal mine IoVT.
Compared to traditional unsupervised generative image enhancement methods, the inclusion of ISP constrains the degrees of freedom in feature mapping, increasing the robustness of style feature enhancement and preventing artifacts. Experimental results demonstrate that the proposed method delivers superior image enhancement in both coal mine and other low-light environments, especially in addressing uneven brightness. Simulated experiments further validate the method's balanced advantage between performance and computational complexity within coal mine IoVT, making it suitable for real-time deployment on edge devices.
In summary, this study addresses the challenges of low-light images impacting monitoring effectiveness in coal mine IoVT and proposes an advanced AI-based image enhancement solution. This method significantly improves monitoring performance, ensuring safer coal mining operations.


%

\bibliographystyle{IEEEtran}

%
%
%
%
%
%
%
%
\bibliography{new}

\begin{thebibliography}{10}
\providecommand{\url}[1]{#1}
\csname url@samestyle\endcsname
\providecommand{\newblock}{\relax}
\providecommand{\bibinfo}[2]{#2}
\providecommand{\BIBentrySTDinterwordspacing}{\spaceskip=0pt\relax}
\providecommand{\BIBentryALTinterwordstretchfactor}{4}
\providecommand{\BIBentryALTinterwordspacing}{\spaceskip=\fontdimen2\font plus
\BIBentryALTinterwordstretchfactor\fontdimen3\font minus
  \fontdimen4\font\relax}
\providecommand{\BIBforeignlanguage}[2]{{%
\expandafter\ifx\csname l@#1\endcsname\relax
\typeout{** WARNING: IEEEtran.bst: No hyphenation pattern has been}%
\typeout{** loaded for the language `#1'. Using the pattern for}%
\typeout{** the default language instead.}%
\else
\language=\csname l@#1\endcsname
\fi
#2}}
\providecommand{\BIBdecl}{\relax}
\BIBdecl

\bibitem{article31}
C.~W. Chen, ``Internet of video things: Next-generation iot with visual
  sensors,'' \emph{IEEE Internet of Things Journal}, vol.~7, no.~8, pp.
  6676--6685, 2020.

\bibitem{article22}
B.~Wu, J.~Wang, B.~Qu, P.~Qi, and Y.~Meng, ``Development, effectiveness, and
  deficiency of china's coal mine safety supervision system,'' \emph{Resources
  Policy}, vol.~82, p. 103524, 2023.

\bibitem{article23}
J.~Zhang, J.~Fu, H.~Hao, G.~Fu, F.~Nie, and W.~Zhang, ``Root causes of coal
  mine accidents: Characteristics of safety culture deficiencies based on
  accident statistics,'' \emph{Process Safety and Environmental Protection},
  vol. 136, pp. 78--91, 2020.

\bibitem{article24}
Y.~Wang, G.~Fu, Q.~Lyu, Y.~Wu, Q.~Jia, X.~Yang, and X.~Li, ``Reform and
  development of coal mine safety in china: An analysis from government
  supervision, technical equipment, and miner education,'' \emph{Resources
  Policy}, vol.~77, p. 102777, 2022.

\bibitem{article30}
M.~Zhao, ``Technology of internet of things technology in the construction of
  smart mine,'' in \emph{2020 3rd International Conference on Artificial
  Intelligence and Big Data (ICAIBD)}.\hskip 1em plus 0.5em minus 0.4em\relax
  IEEE, 2020, pp. 289--292.

\bibitem{article1}
N.~Moroney, ``Local color correction using non-linear masking,'' in \emph{Color
  and Imaging conference}, vol.~8.\hskip 1em plus 0.5em minus 0.4em\relax
  Society of Imaging Science and Technology, 2000, pp. 108--111.

\bibitem{article2}
E.~H. Land, ``The retinex theory of color vision,'' \emph{Scientific american},
  vol. 237, no.~6, pp. 108--129, 1977.

\bibitem{article3}
A.~M. Reza, ``Realization of the contrast limited adaptive histogram
  equalization (clahe) for real-time image enhancement,'' \emph{Journal of VLSI
  signal processing systems for signal, image and video technology}, vol.~38,
  pp. 35--44, 2004.

\bibitem{article6}
J.~Cai, S.~Gu, and L.~Zhang, ``Learning a deep single image contrast enhancer
  from multi-exposure images,'' \emph{IEEE Transactions on Image Processing},
  vol.~27, no.~4, pp. 2049--2062, 2018.

\bibitem{article8}
F.~Lv, F.~Lu, J.~Wu, and C.~Lim, ``Mbllen: Low-light image/video enhancement
  using cnns.'' in \emph{BMVC}, vol. 220, no.~1.\hskip 1em plus 0.5em minus
  0.4em\relax Northumbria University, 2018, p.~4.

\bibitem{article9}
Y.~Zhang, J.~Zhang, and X.~Guo, ``Kindling the darkness: A practical low-light
  image enhancer,'' in \emph{Proceedings of the 27th ACM international
  conference on multimedia}, 2019, pp. 1632--1640.

\bibitem{article4}
K.~G. Lore, A.~Akintayo, and S.~Sarkar, ``Llnet: A deep autoencoder approach to
  natural low-light image enhancement,'' \emph{Pattern Recognition}, vol.~61,
  pp. 650--662, 2017.

\bibitem{article7}
C.~Wei, W.~Wang, W.~Yang, and J.~Liu, ``Deep retinex decomposition for
  low-light enhancement. arxiv 2018,'' \emph{arXiv preprint arXiv:1808.04560}.

\bibitem{article10}
Y.~Jiang, X.~Gong, D.~Liu, Y.~Cheng, C.~Fang, X.~Shen, J.~Yang, P.~Zhou, and
  Z.~Wang, ``Enlightengan: Deep light enhancement without paired supervision,''
  \emph{IEEE transactions on image processing}, vol.~30, pp. 2340--2349, 2021.

\bibitem{article41}
W.~Shi, J.~Cao, Q.~Zhang, Y.~Li, and L.~Xu, ``Edge computing: Vision and
  challenges,'' \emph{IEEE Internet of Things Journal}, vol.~3, no.~5, pp.
  637--646, 2016.

\bibitem{article38}
J.~Wu, R.~Zheng, J.~Jiang, Z.~Tian, W.~Chen, Z.~Wang, F.~R. Yu, and V.~C.
  Leung, ``A lightweight small object detection method based on multi-layer
  coordination federated intelligence for coal mine iovt,'' \emph{IEEE Internet
  of Things Journal}, 2024.

\bibitem{article39}
J.~Wu, R.~Jing, Y.~Bai, Z.~Tian, C.~Wei, S.~Zhang, F.~R. Yu, and V.~C. Leung,
  ``Small insulator defects detection based on multi-scale feature interaction
  transformer for uav-assisted power iovt,'' \emph{IEEE Internet of Things
  Journal}, 2024.

\bibitem{article40}
Y.~Zhao, J.~Wu, W.~Chen, Z.~Wang, Z.~Tian, F.~R. Yu, and V.~C. Leung, ``A small
  object real-time detection method for power line inspection in
  low-illuminance environments,'' \emph{IEEE Transactions on Emerging Topics in
  Computational Intelligence}, 2024.

\bibitem{article13}
W.~Wang, C.~Wei, W.~Yang, and J.~Liu, ``Gladnet: Low-light enhancement network
  with global awareness,'' in \emph{2018 13th IEEE international conference on
  automatic face \& gesture recognition (FG 2018)}.\hskip 1em plus 0.5em minus
  0.4em\relax IEEE, 2018, pp. 751--755.

\bibitem{article5}
L.~Shen, Z.~Yue, F.~Feng, Q.~Chen, S.~Liu, and J.~Ma, ``Msr-net: Low-light
  image enhancement using deep convolutional network,'' \emph{arXiv preprint
  arXiv:1711.02488}, 2017.

\bibitem{article14}
C.~Guo, C.~Li, J.~Guo, C.~C. Loy, J.~Hou, S.~Kwong, and R.~Cong,
  ``Zero-reference deep curve estimation for low-light image enhancement,'' in
  \emph{Proceedings of the IEEE/CVF conference on computer vision and pattern
  recognition}, 2020, pp. 1780--1789.

\bibitem{article11}
W.~Wu, J.~Weng, P.~Zhang, X.~Wang, W.~Yang, and J.~Jiang, ``Uretinex-net:
  Retinex-based deep unfolding network for low-light image enhancement,'' in
  \emph{Proceedings of the IEEE/CVF conference on computer vision and pattern
  recognition}, 2022, pp. 5901--5910.

\bibitem{article33}
W.~Wang, D.~Yan, X.~Wu, W.~He, Z.~Chen, X.~Yuan, and L.~Li, ``Low-light image
  enhancement based on virtual exposure,'' \emph{Signal Processing: Image
  Communication}, vol. 118, p. 117016, 2023.

\bibitem{article34}
Y.~Fan, Y.~Wang, D.~Liang, Y.~Chen, H.~Xie, F.~L. Wang, J.~Li, and M.~Wei,
  ``Low-facenet: Face recognition-driven low-light image enhancement,''
  \emph{IEEE Transactions on Instrumentation and Measurement}, 2024.

\bibitem{article35}
Y.~Cai, H.~Bian, J.~Lin, H.~Wang, R.~Timofte, and Y.~Zhang, ``Retinexformer:
  One-stage retinex-based transformer for low-light image enhancement,'' in
  \emph{Proceedings of the IEEE/CVF International Conference on Computer
  Vision}, 2023, pp. 12\,504--12\,513.

\bibitem{article36}
S.~Yang, M.~Ding, Y.~Wu, Z.~Li, and J.~Zhang, ``Implicit neural representation
  for cooperative low-light image enhancement,'' in \emph{Proceedings of the
  IEEE/CVF international conference on computer vision}, 2023, pp.
  12\,918--12\,927.

\bibitem{article37}
Z.~Fu, Y.~Yang, X.~Tu, Y.~Huang, X.~Ding, and K.-K. Ma, ``Learning a simple
  low-light image enhancer from paired low-light instances,'' in
  \emph{Proceedings of the IEEE/CVF conference on computer vision and pattern
  recognition}, 2023, pp. 22\,252--22\,261.

\bibitem{article15}
Z.~Liang, C.~Li, S.~Zhou, R.~Feng, and C.~C. Loy, ``Iterative prompt learning
  for unsupervised backlit image enhancement,'' in \emph{Proceedings of the
  IEEE/CVF International Conference on Computer Vision}, 2023, pp. 8094--8103.

\bibitem{article16}
M.~Wang, J.~Xing, and Y.~Liu, ``Actionclip: A new paradigm for video action
  recognition,'' \emph{arXiv preprint arXiv:2109.08472}, 2021.

\bibitem{article17}
G.~Li, F.~He, and Z.~Feng, ``A clip-enhanced method for video-language
  understanding,'' \emph{arXiv preprint arXiv:2110.07137}, 2021.

\bibitem{article18}
M.~Tang, Z.~Wang, Z.~Liu, F.~Rao, D.~Li, and X.~Li, ``Clip4caption: Clip for
  video caption,'' in \emph{Proceedings of the 29th ACM International
  Conference on Multimedia}, 2021, pp. 4858--4862.

\bibitem{article21}
L.~Ma, T.~Ma, R.~Liu, X.~Fan, and Z.~Luo, ``Toward fast, flexible, and robust
  low-light image enhancement,'' in \emph{Proceedings of the IEEE/CVF
  conference on computer vision and pattern recognition}, 2022, pp. 5637--5646.

\bibitem{article20}
Z.~Cui, K.~Li, L.~Gu, S.~Su, P.~Gao, Z.~Jiang, Y.~Qiao, and T.~Harada, ``You
  only need 90k parameters to adapt light: a light weight transformer for image
  enhancement and exposure correction,'' \emph{arXiv preprint
  arXiv:2205.14871}, 2022.

\end{thebibliography}
\vspace{-1.5cm}
\begin{IEEEbiography}[{\includegraphics[width=1in,height=1.25in,clip,keepaspectratio]{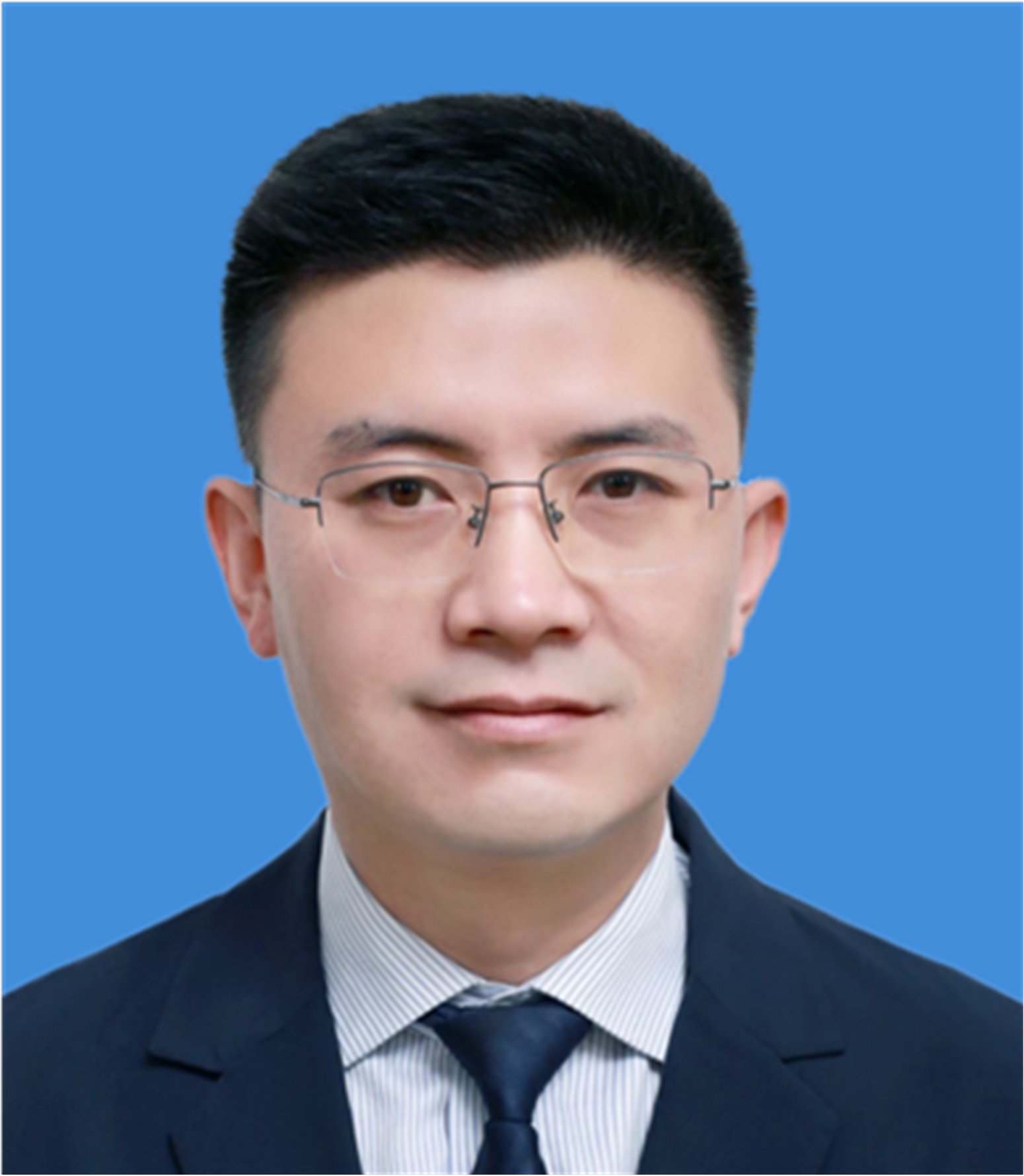}}]{Shuai Wang}received the B.S. degree in automation and the M.S. degree in Mining and Mineral Engineering from Inner Mongolia University of Science and Technology, China, in 2006 and 2015, respectively. He works for the Inner Mongolia Mine Safety Administration. He is currently pursuing the Ph.D. degree with the School of Artificial Intelligence, China University of Mining and Technology (Beijing). His research interests include mine safety monitoring and communication system, machine learning and machine vision.\end{IEEEbiography}
\vspace{-1.0cm}
\begin{IEEEbiography}[{\includegraphics[width=1in,height=1.25in,clip,keepaspectratio]{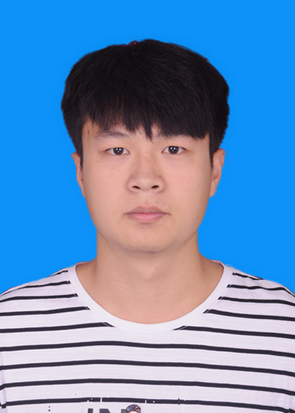}}]{Shihao Zhang} is a Second year graduate student at China University of Mining and Technology(Beijing), and his research focuses on computer vision.\end{IEEEbiography}
\vspace{-1.0cm}
\begin{IEEEbiography}[{\includegraphics[width=1in,height=1.25in,clip,keepaspectratio]{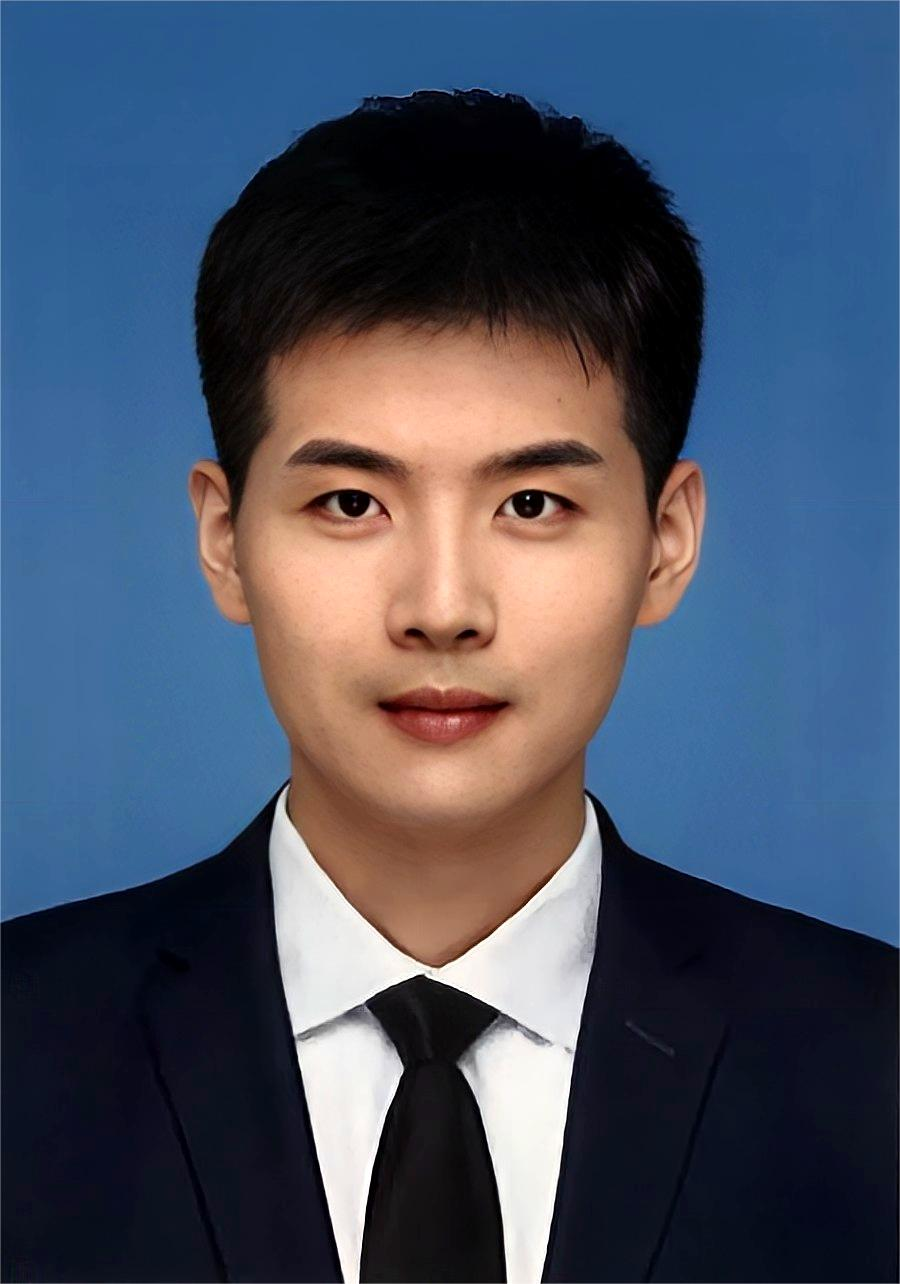}}]{Jiaqi Wu}
	is a third-year doctoral candidate at China University of Mining and Technology(Beijing), and his research focuses on computer vision and computer graphics. He was sent by Chinese Scholarship Council to the University of British Columbia as a joint doctoral student in 2023 to study the Internet of Things and blockchain technology.\end{IEEEbiography}
\vspace{-1.3cm}
\begin{IEEEbiography}[{\includegraphics[width=1in,height=1.25in,clip,keepaspectratio]{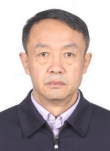}}]{Zijian Tian}
	received the B.S. degree from Xi’an University of Science and Technology, Shanxi,China, and the M.S. degree from China University of Mining and Technology, Beijing, China, in1986 and 1995, respectively, and the Ph.D. degree in communications and information systems from China University of Mining and Technology, Beijing, China, in 2003. In 2003, he joined the School of Mechanical Electronic and Information Engineering, China University of Mining and Technology(Beijing), where he is currently a professor. He is a member of National Expert Group of Safety Production, and is a member and convener of Expert Committee of Information and Automation, a branch of Coal Industry Committee of Technology. His research interests include image processing, analysis of coal mine surveillance video.\end{IEEEbiography}

\vspace{-1.5cm}
\begin{IEEEbiography}[{\includegraphics[width=1in,height=1.25in,clip,keepaspectratio]{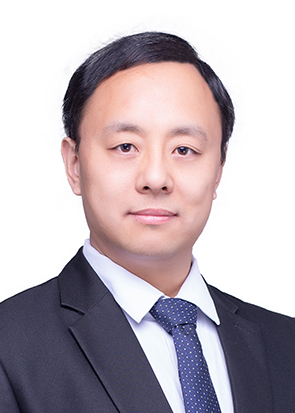}}]{Wei Chen}
	[M’18] received the B.Eng. degree in medical imaging and the M.S. degree in paleontology and stratigraphy from China University of Mining and Technology, Xuzhou, China, in 2001 and 2005, respectively, and the Ph.D. degree in communications and information systems from China University of Mining and Technology, Beijing, China, in 2008. In 2008, he joined the School of Computer Science and Technology, China University of Mining and Technology, where he is currently a professor. He is a member of IEEE, EAI and CCF. He is an associate editor of international journal Array. His research interests include machine learning, image processing, and computer networks.\end{IEEEbiography}
\vspace{-1.3cm}
\begin{IEEEbiography}[{\includegraphics[width=1in,height=1.25in,clip,keepaspectratio]{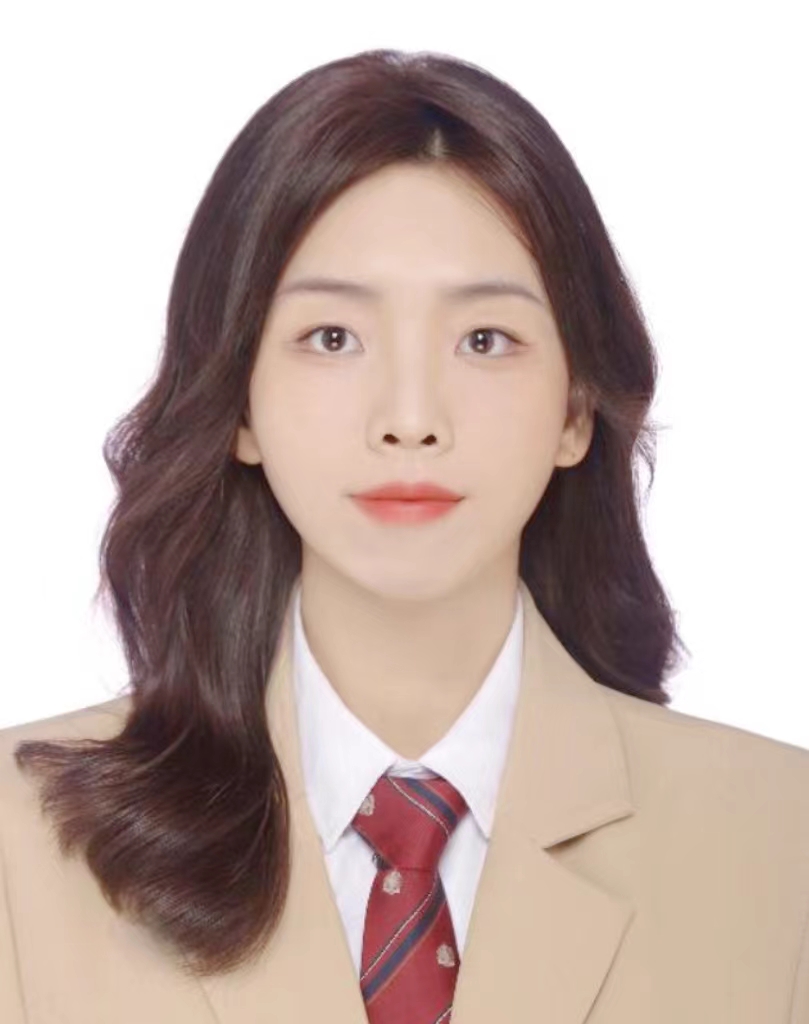}}]{Tongzhu Jin} is a second year masters student at China University of Mining and Technology(Beijing), and her research  interests include security in	blockchain, zero-knowledge proof and data privacy.\end{IEEEbiography}
\vspace{-1.3cm}
\begin{IEEEbiography}[{\includegraphics[width=1in,height=1.25in,clip,keepaspectratio]{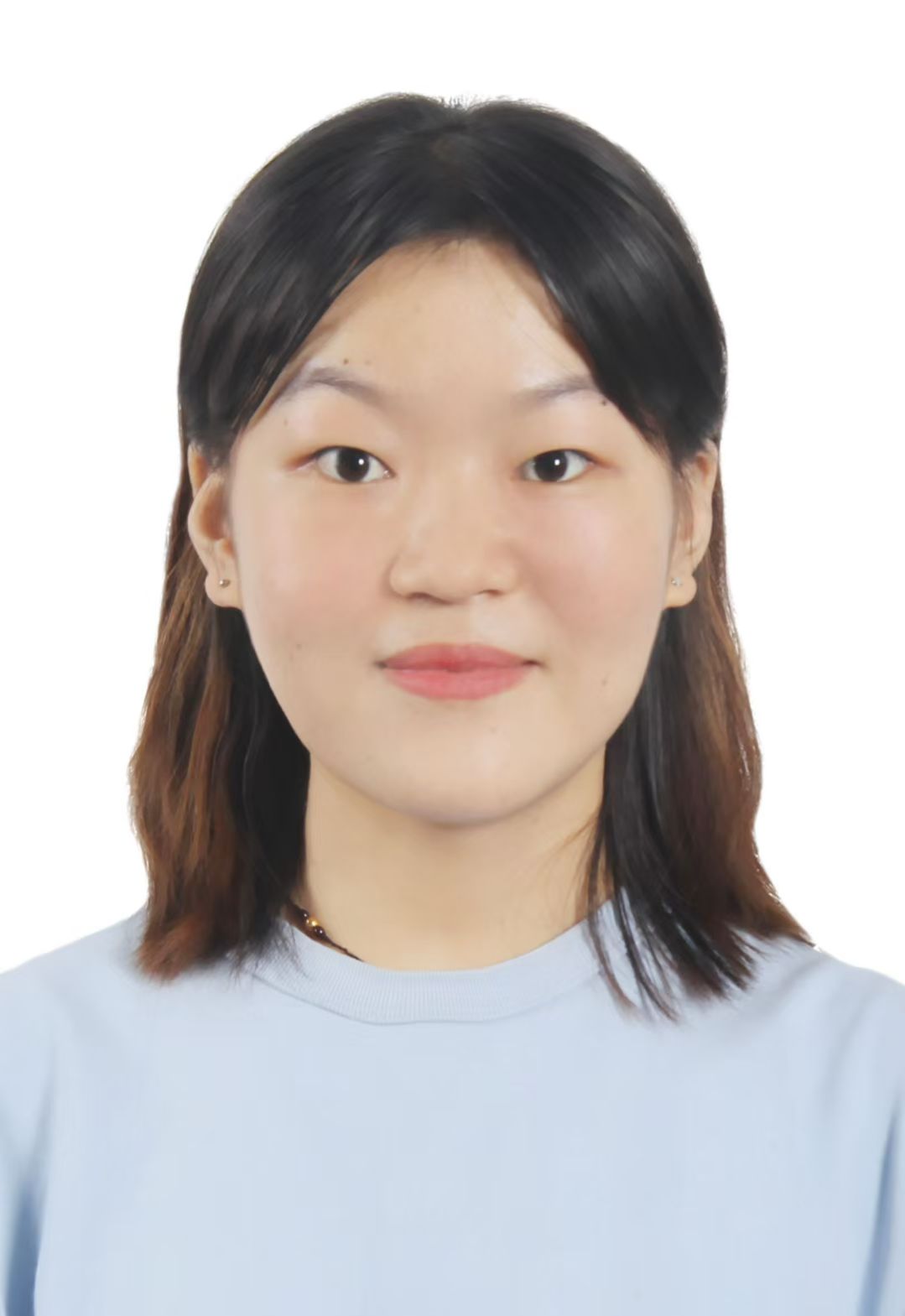}}]{Miaomiao Xue} is a second year masters student at China University of Mining and Technology(Beijing), and her research focuses on machine learning-assisted materials design.\end{IEEEbiography}
\vspace{-1.3cm}

\begin{IEEEbiography}[{\includegraphics[width=1in,height=1.25in,clip,keepaspectratio]{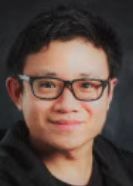}}]{Zehua Wang}
	received Ph.D. degree from The University of British Columbia (UBC), Vancouver in 2016. He is an Adjunct Professor in the Department of Electrical and Computer Engineering at UBC, Vancouver and the CTO at Intellium Technology Inc. He implemented the payment system for commercialized mobile ad-hoc networks with the payment/state channel technology on blockchain. His is interested in protocol and mechanism design with optimization and game theories to improve the efficiency and robustness for communication networks, distributed systems, and social networks.\end{IEEEbiography}
\vspace{-1.3cm}
\begin{IEEEbiography}[{\includegraphics[width=1in,height=1.25in,clip,keepaspectratio]{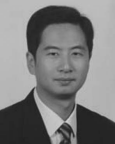}}]{F. Richard Yu}
	received the PhD degree in electrical engineering from the University of British Columbia (UBC) in 2003. His research interests include connected/autonomous vehicles, artificial intelligence, blockchain, and wireless systems. He has been named in the Clarivate’s list of ``Highly Cited Researchers" in computer science since 2019, Standford’s Top 2$\%$ Most Highly Cited Scientist since 2020. He received several Best Paper Awards from some first-tier conferences, Carleton Research Achievement Awards in 2012 and 2021, and the Ontario Early Researcher Award (formerly Premiers Research Excellence Award) in 2011. He is a Board Member the IEEE VTS and the Editor-in-Chief for IEEE VTS Mobile World newsletter. He is a Fellow of the IEEE, Canadian Academy of Engineering (CAE), Engineering Institute of Canada (EIC), and IET. He is a Distinguished Lecturer of IEEE in both VTS and ComSoc.\end{IEEEbiography}
\vspace{-1.3cm}
\begin{IEEEbiography}[{\includegraphics[width=1in,height=1.25in,clip,keepaspectratio]{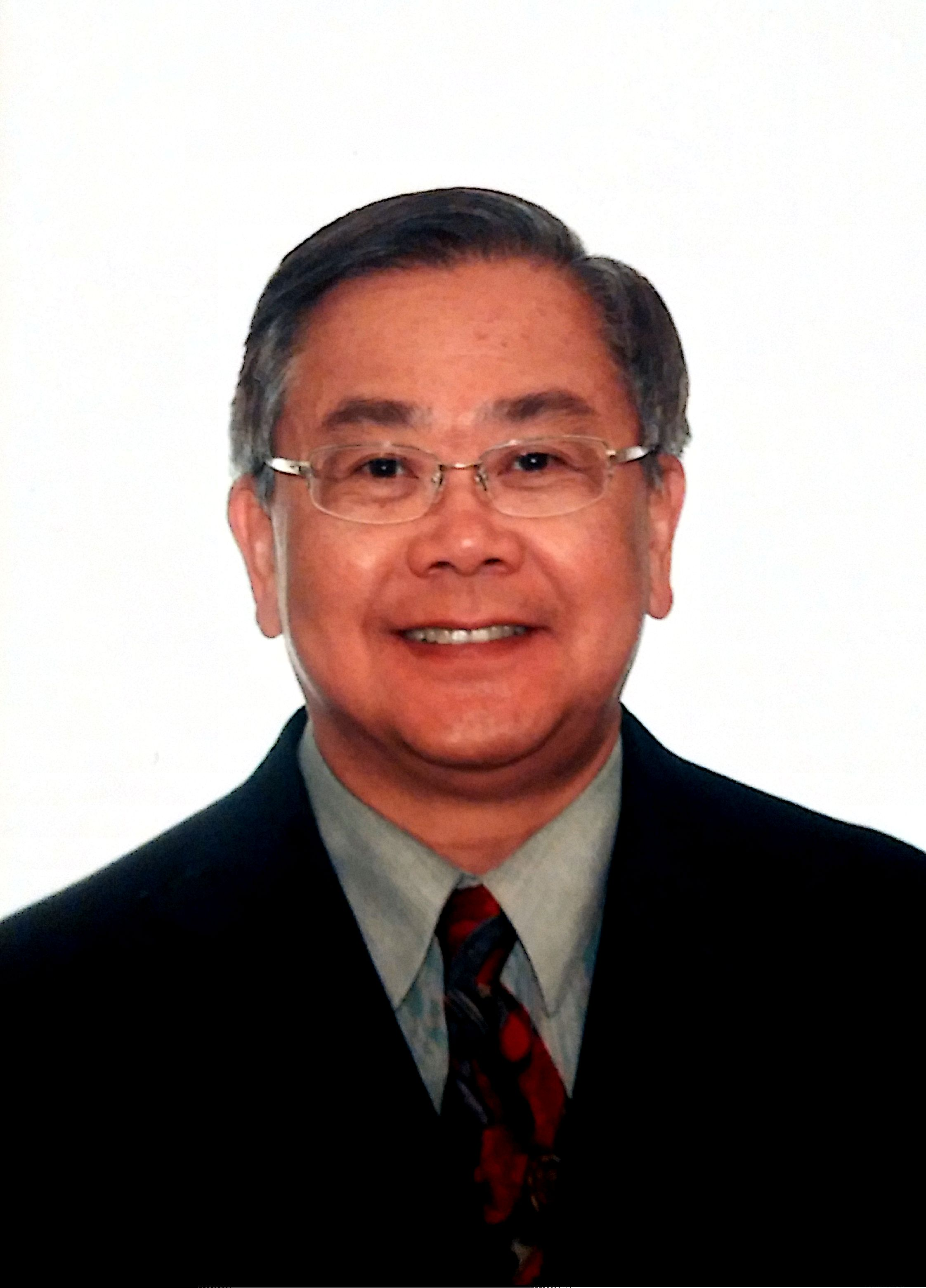}}]{VICTOR C. M. LEUNG}(Life Fellow, IEEE) is the Dean of the Artificial Intelligence Research Institute at Shenzhen MSU-BIT University (SMBU), China, a Distinguished Professor of Computer Science and Software Engineering at Shenzhen University, China, and also an Emeritus Professor of Electrical and Computer Engineering and Director of the Laboratory for Wireless Networks and Mobile Systems at the University of British Columbia (UBC), Canada.  His research is in the broad areas of wireless networks and mobile systems, and he has published widely in these areas. His published works have together attracted more than 60,000 citations. He is named in the current Clarivate Analytics list of “Highly Cited Researchers”. Dr. Leung is serving on the editorial boards of the IEEE Transactions on Green Communications and Networking, IEEE Transactions on Computational Social Systems, and several other journals. He received the 1977 APEBC Gold Medal, 1977-1981 NSERC Postgraduate Scholarships, IEEE Vancouver Section Centennial Award, 2011 UBC Killam Research Prize, 2017 Canadian Award for Telecommunications Research, 2018 IEEE TCGCC Distinguished Technical Achievement Recognition Award, and 2018 ACM MSWiM Reginald Fessenden Award. He co-authored papers that were selected for the 2017 IEEE ComSoc Fred W. Ellersick Prize, 2017 IEEE Systems Journal Best Paper Award, 2018 IEEE CSIM Best Journal Paper Award, and 2019 IEEE TCGCC Best Journal Paper Award.  He is a Life Fellow of IEEE, and a Fellow of the Royal Society of Canada (Academy of Science), Canadian Academy of Engineering, and Engineering Institute of Canada.\end{IEEEbiography}



\end{document}